\documentclass[11pt]{article}
\usepackage{times,amsmath,amssymb,indentfirst,epsfig,longtable}

\begin{document}

\begin{titlepage}
\vspace*{1.2in}
\begin{center}
\hspace*{1.2cm}{\large{\bf Oiling the Wheels of Change}}\\
\hspace*{1.2cm}{\large{\bf The Role of Adaptive Automatic Problem Decomposition in Non--Stationary Environments}}\\
\hspace*{1.2cm}~\\
\hspace*{1.2cm}~\\
\hspace*{1.2cm}{\bf Hussein A. Abbass}\\
\hspace*{1.2cm}{\bf Kumara Sastry}\\
\hspace*{1.2cm}{\bf David E. Goldberg}\\
\hspace*{1.2cm}~\\
\hspace*{1.2cm}~\\
\hspace*{1.2cm}~\\
\hspace*{1.2cm}IlliGAL Report No. 2004020\\
\hspace*{1.2cm}May, 2004\\
\vspace*{6cm} {\large
\hspace*{1.2cm}Illinois Genetic Algorithms Laboratory (IlliGAL)\\
\hspace*{1.2cm}Department of General Engineering\\
\hspace*{1.2cm}University of Illinois at Urbana-Champaign \\
\hspace*{1.2cm}117 Transportation Building \\
\hspace*{1.2cm}104 S. Mathews Avenue, Urbana, IL 61801 \\}
\end{center}
\end{titlepage}

\title{Oiling the Wheels of Change:\\The Role of Adaptive Automatic Problem Decomposition in Non--Stationary Environments}

\author{Hussein A. Abbass\thanks{Artificial Life and Adaptive Robotics Laboratory,
School of Information Technology and Electrical Engineering,
University of New South Wales, Australian Defence Force Academy,
Canberra, ACT 2600, Australia. h.abbass@adfa.edu.au},   Kumara
Sastry\thanks{Illinois Genetic Algorithms Laboratory, University
of Illinois, 117 Transportation Building, 104 S. Mathews Av.
Urbana, IL 61801, deg@illigal.ge.uiuc.edu}, and David
Goldberg\thanks{Illinois Genetic Algorithms Laboratory, University
of Illinois, 117 Transportation Building, 104 S. Mathews Av.
Urbana, IL 61801, deg@illigal.ge.uiuc.edu}}

\date{}
\maketitle

\begin{abstract}

{\em Genetic algorithms} (GAs) that solve hard problems quickly,
reliably and accurately are called {\em competent\/} GAs. When the
fitness landscape of a problem changes overtime, the problem is
called non--stationary, dynamic or time--variant problem. This
paper investigates the use of competent GAs for optimizing
non--stationary optimization problems. More specifically, we use
an information theoretic approach based on the minimum description
length principle to adaptively identify regularities and
substructures that can be exploited to respond quickly to changes
in the environment. We also develop a special type of problems
with bounded difficulties to test non--stationary optimization
problems. The results provide new insights into non-stationary
optimization problems and show that a search algorithm which
automatically identifies and exploits possible decompositions is
more robust and responds quickly to changes than a simple genetic
algorithm.
\end{abstract}

\section{Introduction}

Real--world problems are rarely static. Problems change overtime,
a factor compounded by the fact that environments under which they
function are also in a constant state of flux. Although
significant advances have been made in the development and design
of genetic and evolutionary algorithms
\cite{Foge95b,Gold02,Rechenberg1,Rechenberg2}, only a few have
accounted for the changing nature of the problems themselves
\cite{Bran01}. Resolving problems from scratch every time a change
occurs is neither practical nor feasible, and is tantamount to
re-inventing the wheel every time a problem with the wheel occurs.
This aspect of problem-solving is especially pertinent where
change is so frequent that re-solving the original problem can
never be appropriate. \\

We hypothesize that to solve non-stationary problems efficiently,
previously encountered solutions can be used to extract structural
knowledge about the problem in hand. Identifying important
regularities and sub--structures in a problem can help in
responding quickly and tracking optima when the environment
changes. A class of evolutionary algorithms that automatically
discover the problem decomposition is known as {\em competent}
genetic algorithms
\cite{Gold99,Larr02,Balu94,Hari98,Hari99,Karg95,Muh96,Mun99,Peli02}.
In essence, competent genetic algorithms automatically and
adaptively identify important sub--structures of an underlying
search problem and use them to efficiently explore the search
space. The aim of this paper is to explore the advantages of using
a candidate of these methods to examine our hypothesis. \\

More specifically, we use the {\em extended compact genetic
algorithm} (ecGA) \cite{Hari99} as a candidate of probabilistic
model--building GAs. In these types of GAs, the variation
operators are replaced by building and sampling a probabilistic
model of promising solutions. In ecGA, the probabilistic model is
based on the information theoretic measure known as the minimum
description length principle \cite{Riss78,Riss89,Riss96}. The
structure and the probabilities of the decomposition model is
manipulated when the environment changes to speed-up the response
of the solver to the changes. Similar to other studies on using
genetic and evolutionary algorithms on non-stationary problems, we
assume that the new solutions are related to the old ones and that
the changes are bounded. Specifically, we incorporate bounded
changes to both the problem structure and the fitness landscape.
It should be noted that if the environment changes either
unboundedly or randomly, on average no method will outperform
restarting the solver from scratch every time a change
occurs. \\

The structure of the paper is as follows: in the next section, we
will present a brief review to the background materials relevant
to this paper. We will then review ecGA followed by the different
methods we use for dynamic optimization in this paper. We then
present the experimental setup, results, and discussions.  \\

\section{Background Materials}

In this section, we present a brief overview to previous work on
evolutionary computation methods for dynamic environments and
Adaptive automatic decomposition approaches.\\

\subsection{Dynamic Environments}

To date, there have been three main evolutionary approaches to
solve optimization problems in changing environments. These
approaches are: diversity control, memory-based, and
multi-population methods. We will present a brief overview to this
literature here and refer the reader to \cite{Bran01} for a more
detailed review to this large growing field. \\

Diversity has been a focal point of many recent work in enhancing
the adaptiveness of evolutionary methods for dynamic optimization
problems. Diversity is controlled in two ways; either by
increasing the diversity whenever a change is detected or
maintaining high diversity all over the evolutionary run. Examples
of the former include the hyper--mutation method \cite{Cobb}, the
variable local search technique \cite{Vava97} and other methods in
\cite{Bier99} and \cite{Lin97}. The main methods in the latter
group include Redundancy
\cite{Gold89,Gold87,Coll97,Dag95,Wine00,Urse00}, random immigrants
\cite{Gref92}, Aging \cite{Ghos98}, and the Thermodynamical
Genetic Algorithms \cite{Mori96,Mori97}.\\

Memory-based approaches attract much attention in the literature.
Two main types exist, implicit and explicit memories. In implicit
memory \cite{Gold87,Hada97,Dasg92b}, a redundant representation is
used as a means for memory. In explicit memories \cite{Mori97},
specific information, which may include solutions, get stored and
retrieved when needed by the evolutionary mechanism.\\

The third class of approaches depends on speciation and
multi-populations. Sub--populations are maintained and each
becomes specialized on a part of the search space. This
facilitates the process of tracking the optima as they move. An
example in this group is the Self-organizing-scouts method
\cite{Bran01}.\\

In all previous work - diversity control, memory-based, and
multi-population methods - the performance of different techniques
may vary by the manner in which the environment changes
\cite{Bran01}. Branke \cite{Bran01} attempted to classify
different types of dynamics to gain an insight of the level of
difficulties in dynamic optimization problems. A major research
question here is what does make a dynamic optimization problem
hard to solve by evolutionary methods? Another equally important
question is whether by learning some decomposition of the problem,
can it help in responding quickly to a change in the environment
assuming that this decomposition is not affected by this change?\\

\subsection{Adaptive Automatic Decomposition}

One of the key challenges in the area of genetic and evolutionary
algorithms is the systematic design of genetic operators with
demonstrated scalability. Based on Holland's \cite{Holl75} notion
of building blocks, Goldberg \cite{Gold02,Gold91,Gold92} proposed
a design--decomposition theory for designing effective GAs. The
theory establishes the identification of suitable substructures or
decompositions (also referred to as linkage) and ensuring
efficient exchange of these substructures as a challenging task in
designing competent GAs. The design--decomposition theory not only
provides an insight into what makes a problem hard for GAs, but
also has resulted in many {\em competent\/} GA designs. In
essence, competent GAs successfully solve problems with bounded
difficulties in a polynomial (sometimes sub--quadratic) number of
function evaluations \cite{Gold02}. A key element of competent GAs
is a mechanism to automatically identify important substructures
of the underlying search problem. Depending on the mechanism used
to discover the problem decomposition, competent genetic
algorithms can be classified into three broad categories:\\

\begin{description}

\item[{\bf Perturbation techniques}] include the messy genetic
algorithm \cite{Gold89a}, fast messy genetic algorithm
\cite{Gold93,Karg95}, gene expression messy genetic algorithm
\cite{Karg96}, linkage identification by nonlinearity check
genetic algorithm, and linkage identification by monotonicity
detection genetic algorithm \cite{Mun99}, and dependency structure
matrix driven genetic algorithm \cite{Yu03}, and linkage
identification by limited probing \cite{Heck03}.\\

\item[{\bf Linkage adaptation techniques}] such as linkage
learning GA \cite{Hari97a,Chen02,Chen04}.\\

\item[{\bf Probabilistic model building techniques}]
\cite{Peli02,Peli02a,Larr02} such as population-based incremental
learning \cite{Balu94}, the bivariate marginal distribution
algorithm \cite{Peli99}, the extended compact GA (ecGA)
\cite{Hari99}, iterated distribution estimation algorithm
\cite{Bosm99}, Bayesian optimization algorithm (BOA)
\cite{Peli00}.\\

\end{description}

A more detailed survey of various problem-decomposition mechanisms
(or genetic linkage learning) are discussed elsewhere and the
references therein \cite{Chen04}.\\

Despite the success of competent GAs in solving stationary search
problems, they have not been used to solve non-stationary problems
apart from a preliminary study by \cite{Sing02}. The aim of this
paper is twofold: first, to examine the performance of ecGA in
terms of its response rate, as an example of a competent GA that
automatically decomposes and identifies substructures in
non--stationary problems; and second, to test the method on
problems with bounded difficulties. Our conjecture is that by
having a mechanism which focuses on identifying the important
substructures (or building blocks) is beneficial for dynamic
optimization problems as well. Furthermore, the
problem-decomposition information serves as a way to store past
information which could be used and manipulated to respond faster
to changes in the environment. \\

\section{The Extended Compact Genetic Algorithm}

The {\em extended compact genetic algorithm} \cite{Hari99} is a
probabilistic model building genetic algorithm which replaces
traditional variation operators of genetic and evolutionary
algorithms by building a probabilistic model of promising
solutions and sampling the model to generate new candidate
solutions. Harik \cite{Hari99} studied the problem of linkage
learning and proposed a conjecture that linkage learning is
equivalent to a good model that learns the structure underlying a
set of genotypes. Being focused on probabilistic models, Harik
focused on probabilistic models to learn linkage. In the ecGA
method, he proposed the use of the {\it minimum description
length} (MDL) principle \cite{Riss78,Riss89,Riss96} to compress
good genotypes into partitions that include the shortest possible
representations. The MDL measure is a tradeoff between two
complexity measures. The first is a measure of information content
in a population which Harik calls ``compressed population
complexity'' while the second is a measure of the size of the
model which Harik calls ``model complexity''.\\

The compressed population complexity measure is a statistical
complexity measure based on the well--known information-theoretic
approach of Shannon's entropy \cite{shannon48}. Shannon's entropy
$E(\chi_I)$ of the population assumes that each partition of
variables $\chi_I$ is a random variable with probability $p_{i}$.
The measure is given by
\begin{equation}
E(\chi_I)=-\ C \sum_{i}^{\sigma} p_{i} \log p_{i}
\end{equation}
where $C$ is a constant related to the base chosen to express the
logarithm and $\sigma$ is the number of all possible bit sequences
for the variables belonging to partition $\chi_I$; that is, if the
cardinality of $\chi_I$ is ${\nu_I}$, $\sigma = 2 ^{\nu_I}$. This
measures the amount of disorder associated within a population
under a decomposition scheme. Equivalently, it can be seen as the
amount of information content presents in the population under a
specific partition scheme. The compressed population complexity is
a scaled version of the entropy as follows

\begin{equation}
Compressed \ \ Population \ \ Complexity \ \ = N \sum_I E(\chi_I)
\end{equation}

The second complexity measure is associated with the model itself,
which measures the complexity of the model in terms of its size as
follows:

\begin{equation}
Model \ \ Complexity \ \ = \log(N+1) \left( 2^{\nu_I} - 1 \right)
\end{equation}

The MDL measure is the sum of the compressed population complexity
and the model complexity as follows

\begin{equation}
MDL \ \ = N \sum_I \left( -\ C \sum_{i}^{\sigma} p_{i} \log p_{i}
\right) + \log(N) 2^{\nu_I}
\end{equation}

The ecGA method can be summarized in the following steps:

\begin{enumerate}
\item Initialize the population at random with $n$ individuals;

\item Evaluate all individuals in the population;

\item Use tournament selection without replacement to select $n$
individuals;

\item Use the MDL measure to recursively partition the variables
until the measure increases;

\item Use the partition to shuffle the building blocks (building
block--wise crossover) to generate a new population of $n$
individuals;

\item If the termination condition is not satisfied, go to 2;
otherwise stop.

\end{enumerate}

\section{Methods}

In this section, we present two variations of the ecGA algorithm
for dynamic environments. We assume in this paper that we have a
mechanism to detect the change in the environment. Detecting a
change in the environment can be done in several ways including:
(1) re--evaluating a number of previous solutions; and (2)
monitoring statistical measures such as the average fitness of the
population \cite{Bran01}. The focus of this paper is not, however,
on how to detect a change in the environment; therefore, we assume
that we can simply detect it. The modified ecGA algorithm for
dynamic environments works as follows:

\begin{enumerate}
\item Initialize the population at random with $n$ individuals;

\item If a change in the environment is being detected, do:

\begin{enumerate}
\item Re--initialize the population at random with $n$
individuals;

\item Evaluate all individuals in the population;

\item Use tournament selection without replacement to select $n$
individuals;

\item Use the last found partition to shuffle the building blocks
(building block--wise crossover) to generate a new population of
$n$ individuals;

\end{enumerate}

\item Evaluate all individuals in the population;

\item Use tournament selection without replacement to select $n$
individuals;

\item Use the MDL measure to recursively partition the variables
until the measure increases;

\item Use the partition to shuffle the building blocks (building
block--wise crossover) to generate a new population of $n$
individuals;

\item If the termination condition is not satisfied, go to 2;
otherwise stop.

\end{enumerate}

We will call the previous version dcGA(1). In this version, once a
change is detected, a new population is generated at random,
followed by selection and crossover using the last generated
model. The method then continues with the new population. In the
second version, dcGA(2), the last learnt model is not used to bias
the re--start mechanism where the steps of selection and crossover
that are carried out on the new randomly generated population are
ignored. Both versions can be seen as a re--start approach, where
the first instance uses the last learnt model after the re--start,
while the second does not. In ecGA, the model is re-built from
scratch in every generation. This has the advantage of recovering
from possible problems that may
exist from the use of a hill--climber in learning the model. \\

Kargupta \cite{Karg95} has shown that problems with bounded
complexity can be solved in a polynomial time ``provided that
there exists an appropriate measure that can correctly detect the
good relations''. M$\ddot{u}$hlenbein \cite{Muh92} showed that
order--k functions with length $l$ are solvable in $O(l^k
\log(l))$ using a $(1+1)ES$. Goldberg et. al. \cite{Gold93}
achieved $O(l^2)$ complexity using the fast messy genetic
algorithms. Pelikan \cite{Peli02} provided a complexity of
$O(n^{1.65})$ using BOA. Sastry and Goldberg has shown that the
convergence time for ecGA follows the relation derived by
M$\ddot{u}$hlenbein and Voosen \cite{Mueh93} for breeder GAs,
where the convergence time is equal to $\frac{\pi \sqrt(l)}{I}$,
where $I$ is the selection intensity and $l$ is the number of bits
in the chromosome.\\

In a changing environment, let us assume a chromosome with $BB$
building blocks each of order $k$ bits, $l=k \times BB$. The ecGa
will behave according to the previous complexity equation to build
a correct decomposition model. If the environment does not affect
the decomposition but only affects the peaks within building
blocks, a complete enumeration of all possible solutions within
each building block would have a time complexity of
$\Theta(m.2^k)$ to get to the new optima. The notation $\Theta$
represents lower and upper bound (tight) complexity. This is not
very expensive. Assume a 5 bit building block replicated 100 times
(a 500 bits problem); the cost of tracking the optima when the
decomposition does not change would be $2^5 \times 100 = 3200$
objective evaluations. This cost is less than what the experiments
will show because the algorithm is designed to handle the general
case that the decomposition may also change rather than the very
specific case of fixed decomposition. \\

We compare the results against a similar genetic algorithm except
that the linkage learning based crossover operator in ecGA is
replaced with a uniform crossover operator. We call this algorithm
uGA to emphasize its use of uniform crossover with genetic
algorithms. In the following section, we will present the
experiments and the test functions used to test the proposed
method. \\

\section{Experiments}

\subsection{Test Functions}

%Goldberg \cite{Gold87b} studied the kind of deception and linkage
%in problems of bounded difficulty  that can mislead a simple
%genetic algorithm. He constructed the {\it minimal deceptive
%problem}, a 2--bit problem that can mislead a GA, and showed that
%it is more likely to solve the problem using a simple GA when the
%bits are close to each other than when the order is loose and the
%bits are separated by a long string.

A special class of problems that represent a challenge to GAs
methods is known as ``problems of bounded difficulty''. These
problems are characterized by two main features: they are
additively decomposable and separable functions, and uniformly
scaled. A function $f(X) = f(x_0, \dots, x_i, \dots, x_n)$ is said
to be additively decomposable and separable iff there exists a
partition of $<\chi_j>_{j=1}^m$ such that $\chi_j \ne \phi, j = 1,
\dots, m$, $\chi_j \bigcap \chi_k = \phi, j \ne k$, and
$\bigcup_{j=1}^m \chi_j = X$. Under this partition scheme, the
function $f(X)$ can be rewritten as
\[ f(X) = \sum_{j=1}^m f_j(\chi_j) \]
The function is said to be uniformly scaled if all $f(\chi_j)$ are
derived from the same class of functions. There are no assumptions
on each $f$; each can be a multimodal function and can take any
function form. Problems of bounded difficulty have been studied
widely because they can provide an easy to analyze test functions
which challenge the dynamics of simple genetic algorithms. We will
define the order of difficulty for such a problem as $k = \max_j
|\chi_j| << n$, with $|.|$ represents the cardinality of the set.
Solving a problem with bounded difficulty becomes easy once the
variables can be correctly separated into the right partitions; at
which point, a complete enumeration of all possible solutions for
each partition is sufficient to find the global optimal solution.
Here we assume that the cardinality of each partition is small and
is much smaller than the length of the solution vector. However,
in the absence of the value of $k$ and any knowledge of which
variable belongs to which partition, the problem can be tough.
Examples of problems with bounded difficulties include the Ising
problem \cite{Hoy01,Hoy02}, trap functions
\cite{Ack87,Deb93,Thie93}, and functions which incorporate the
notion of multimodality, hierarchy, crosstalk and deception
\cite{Gold02}. These test problems, despite being easy to
understand, incorporates many of the essential difficulties for
linkage identification.\\

\subsection{Experimental Design}

We repeated each experiment 30 times with different seeds. All
results are presented for the average performance over the 30
runs. The population size is fixed to 5000 in all experiments. The
population size is chosen large enough to provide enough samples
for the probabilistic model to learn the structure and to provide
enough diversity for uGA. Termination occurs when the algorithm
reaches the maximum number of generations of 100. We assume that
the environment changes between generations and the changes in the
environment are assumed to be cyclic, where we tested two cycles
of length 5 and 10 generations respectively. The crossover
probability is 1, and the tournament size is set to 16 in all
experiments based on Harik's default values. \\

\subsection{Experiment 1}

The method is tested using dynamic versions of three trap
functions. Trap functions were introduced by Ackley \cite{Ack87}
and subsequently analyzed in details by others
\cite{Deb93,Gold02,Thie93}. A trap function is defined as follows
\begin{equation}
trap_k = \left\{ \begin{array}{ll} high & \text{if } u=k \\ low -
u * \frac{low}{k-1} & otherwise \end{array} \right.
\end{equation}
where, $low$ and $high$ are scalars, $u$ is the number of 1s in
the string, and $k$ is the order of the trap function. In this
paper, we choose $low=k$, $high=k+1$. \\

\begin{figure} [h!]
\begin{center}
 \epsfig{figure=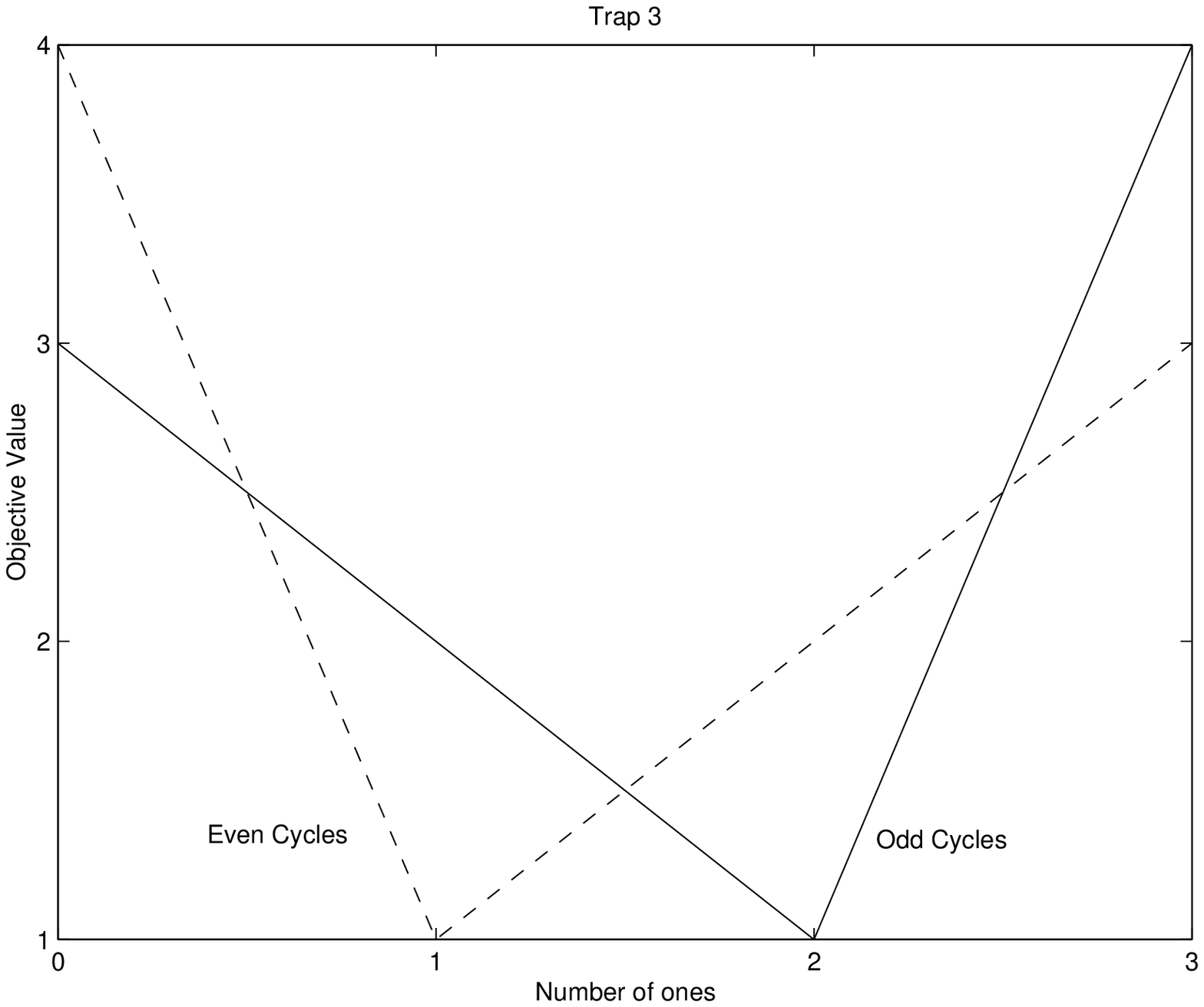, width=2in, height=2in}
 \epsfig{figure=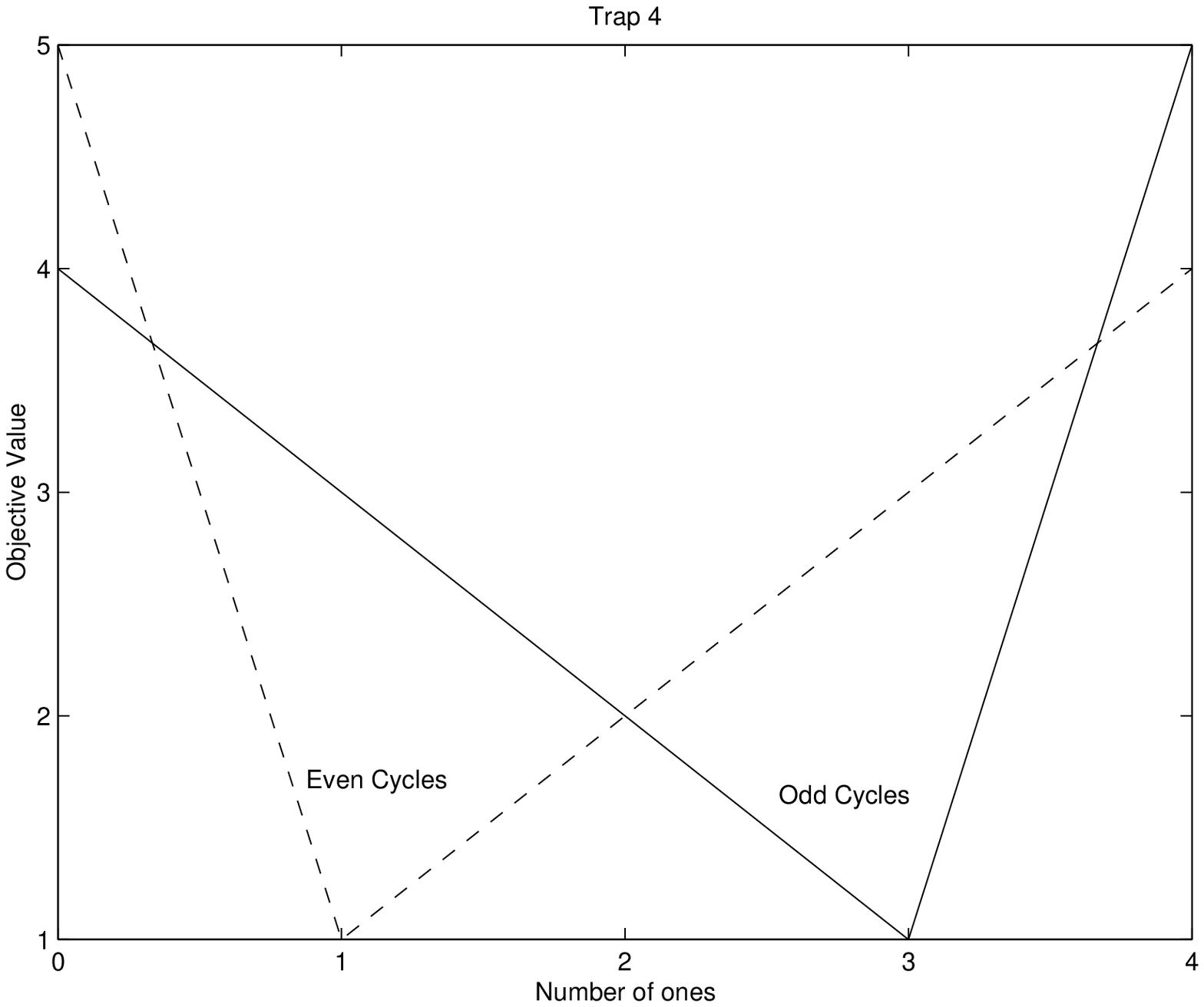, width=2in, height=2in}
 \epsfig{figure=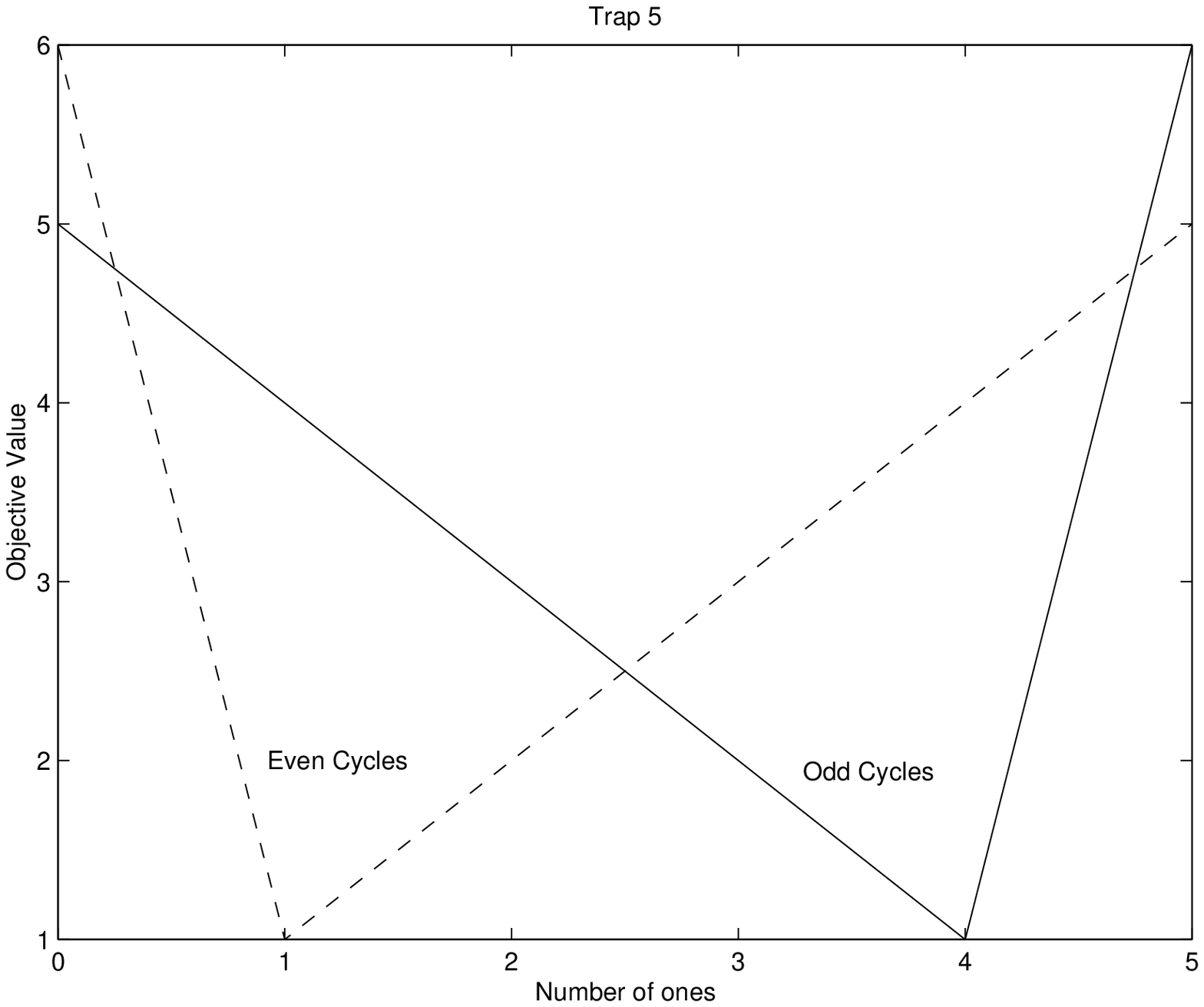, width=2in, height=2in}
 \caption{The Trap functions in a changing environment. (Left) trap--3; (middle) trap--4; (right) trap--5.}\label{ftrap1}
\end{center}
\end{figure}

In the initial set of experiments, we tested the method using
traps of order 3, 4, and 5. Figure~\ref{ftrap1} depicts a
graphical representation of the traps and how they change. In odd
cycles, the global optimum is when all variables are 1's, while in
even cycles and at time 0, the global optimum is when all
variables are 0's.\\

We tested the methods with 5, 10, 15, and 20 building blocks. If
we denote the number of building blocks by $BB$, then the optimal
solution for each problem would be at $BB * (k+1)$. For example,
with 20 building blocks in trap--5, the optimal solution has an
objective value of 120 regardless of the change in the
environment. The environment in this first experiment does not
actually change the value of the optimal solution but severely
changes the value of the decision variables. The change is severe
as the optimal solutions isolates between two points separated
with the maximum possible hamming distance in the hamming subspace
defined by each trap. \\

\begin{figure} [h!]
\begin{center}
 \epsfig{figure=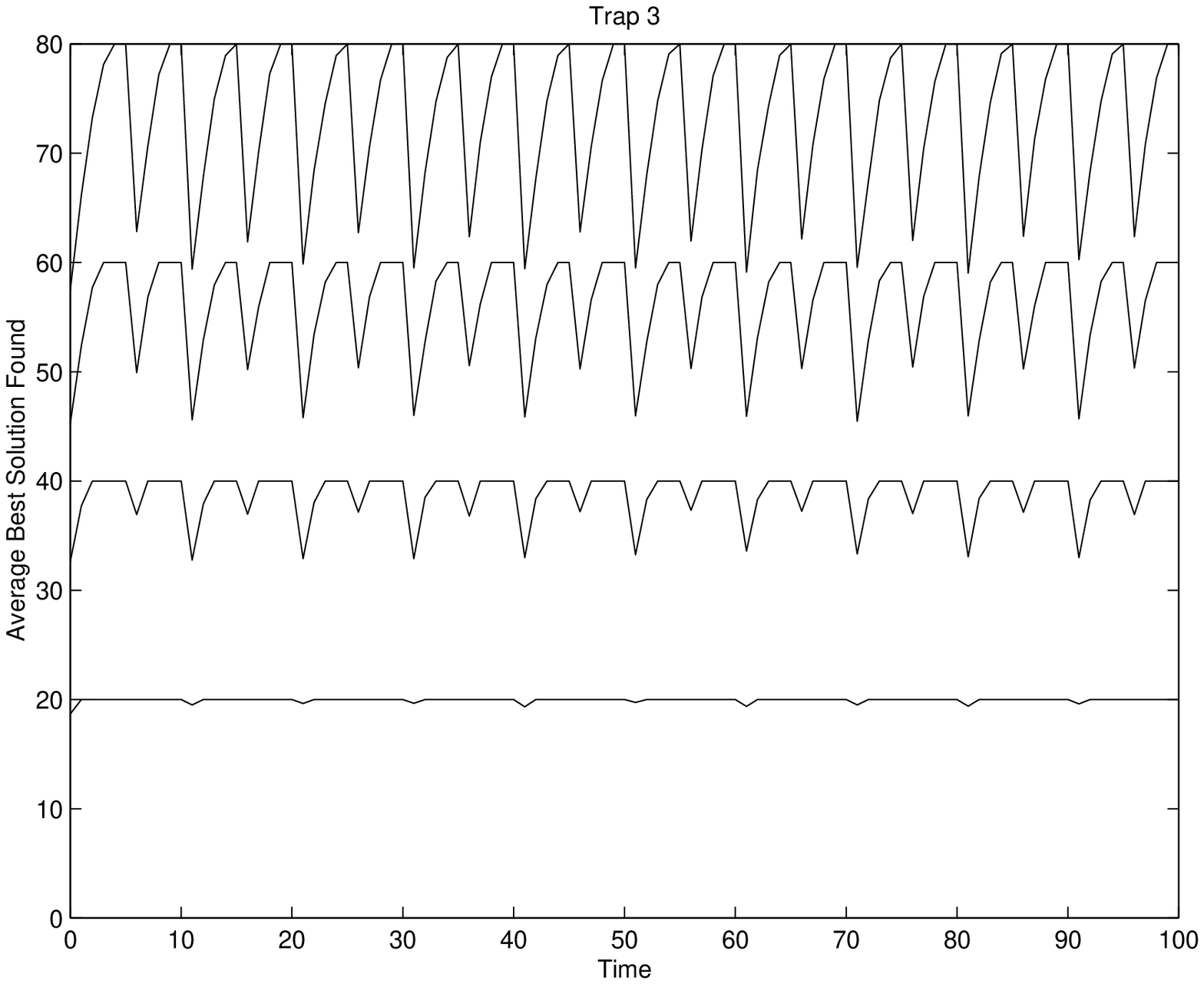,  width=2in, height=2in}
 \epsfig{figure=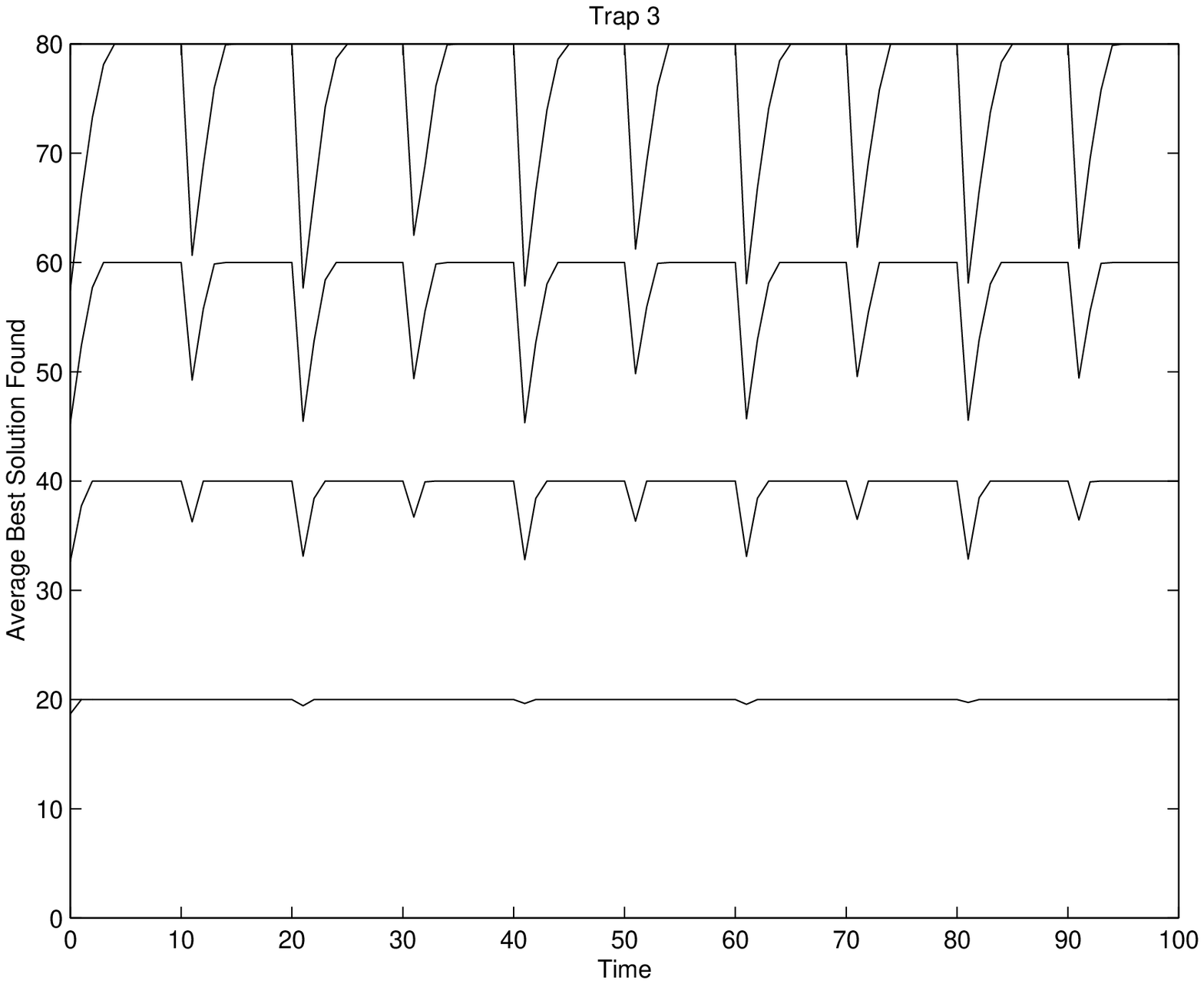,  width=2in, height=2in}

 \epsfig{figure=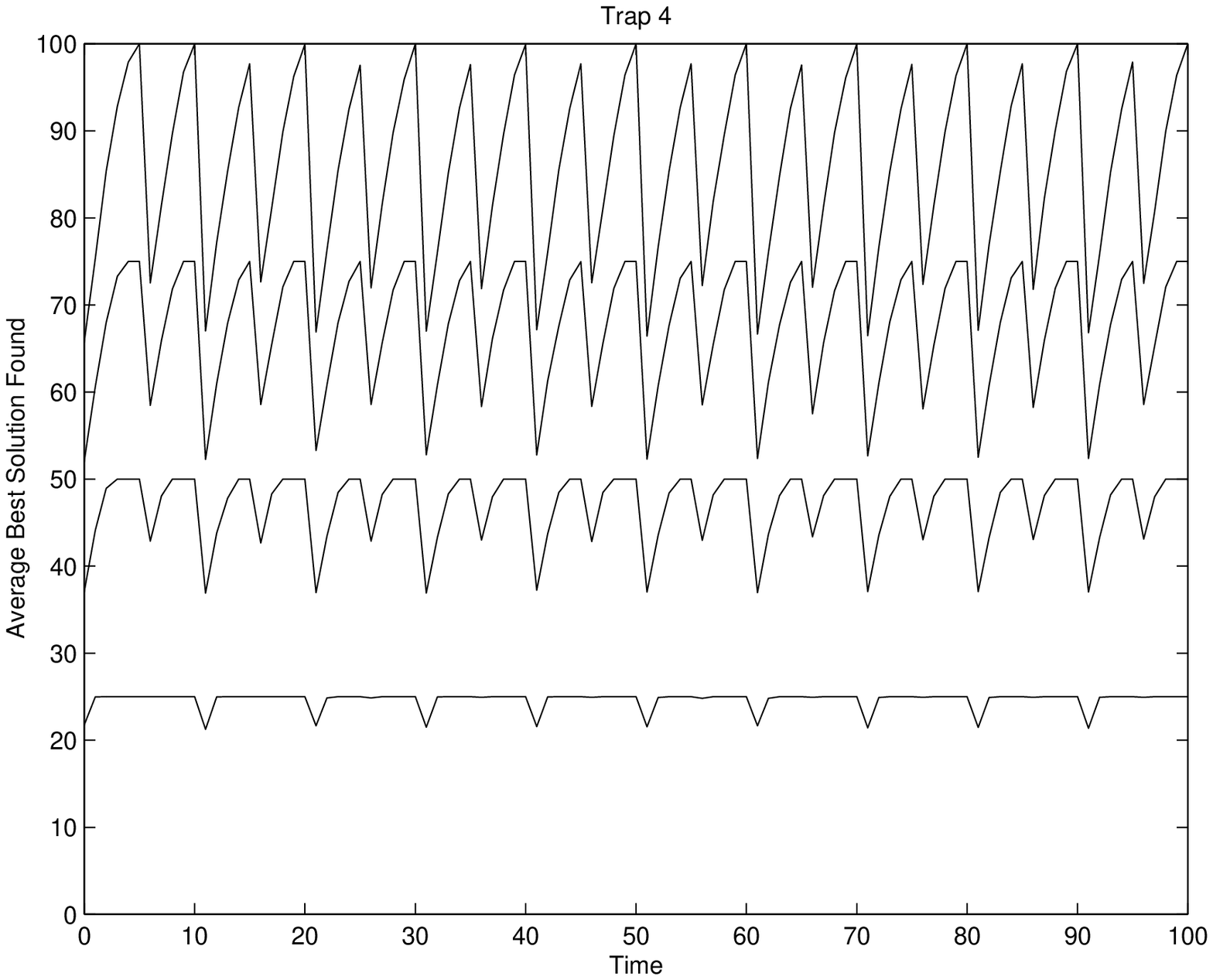,  width=2in, height=2in}
 \epsfig{figure=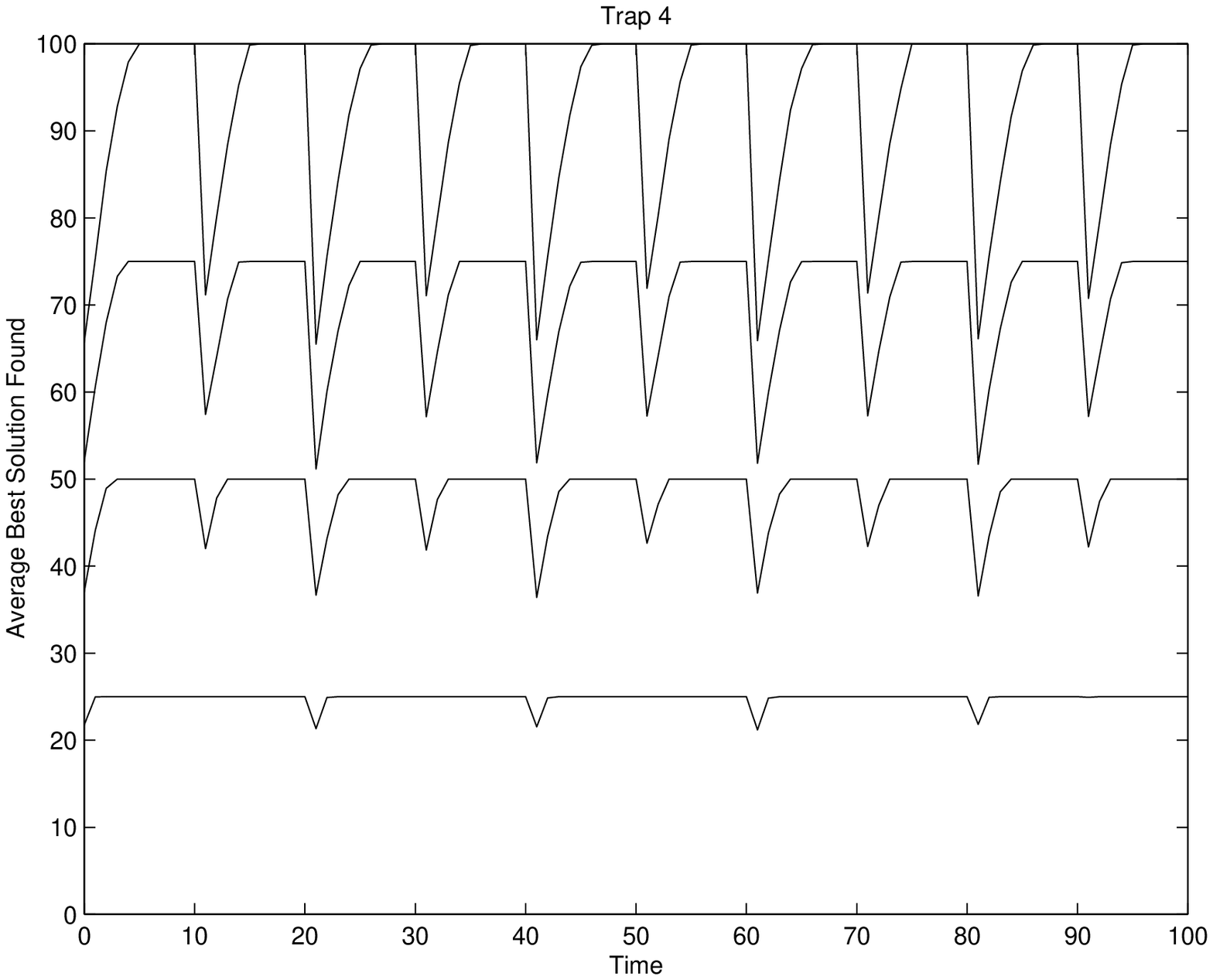,  width=2in, height=2in}

 \epsfig{figure=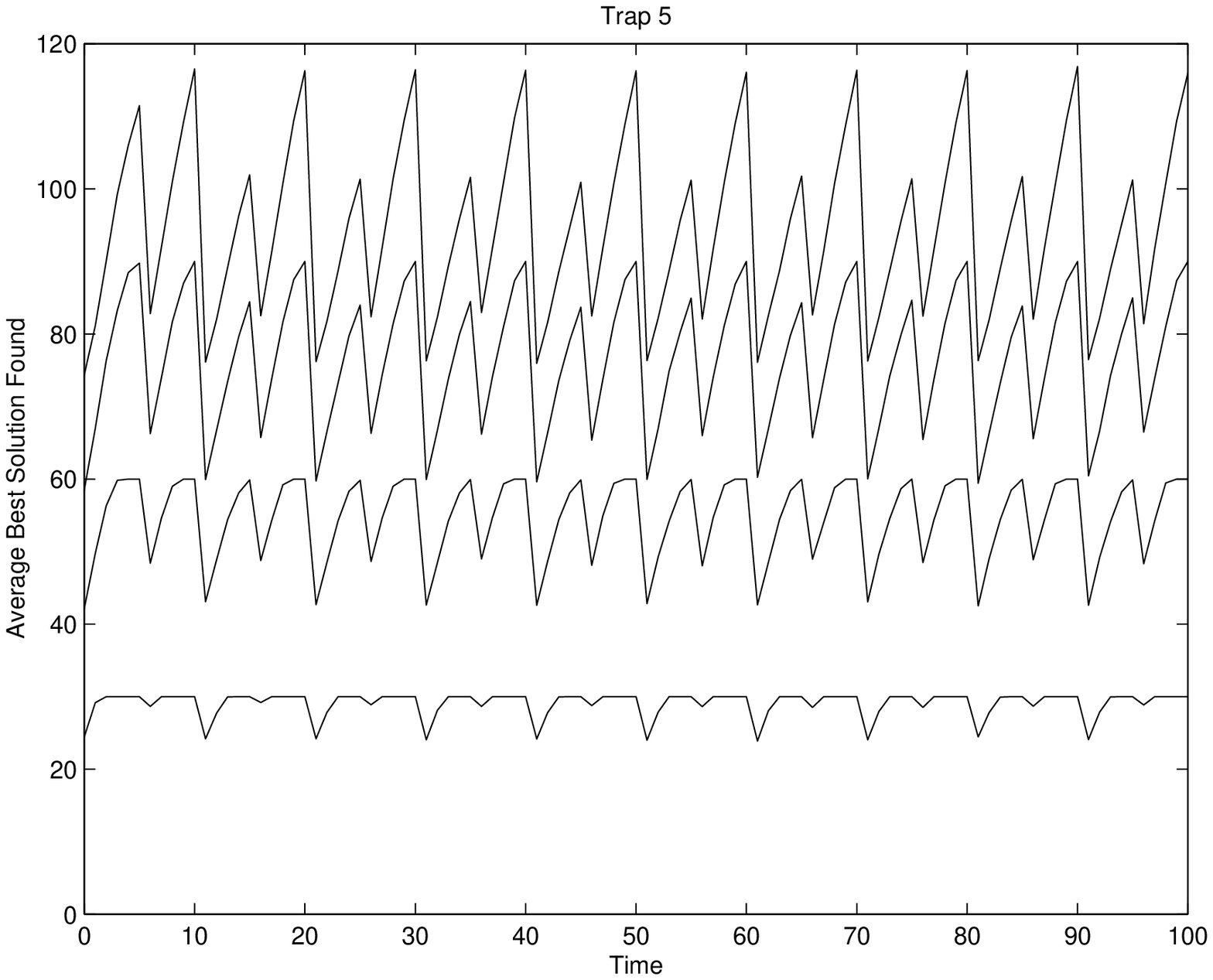,  width=2in, height=2in}
 \epsfig{figure=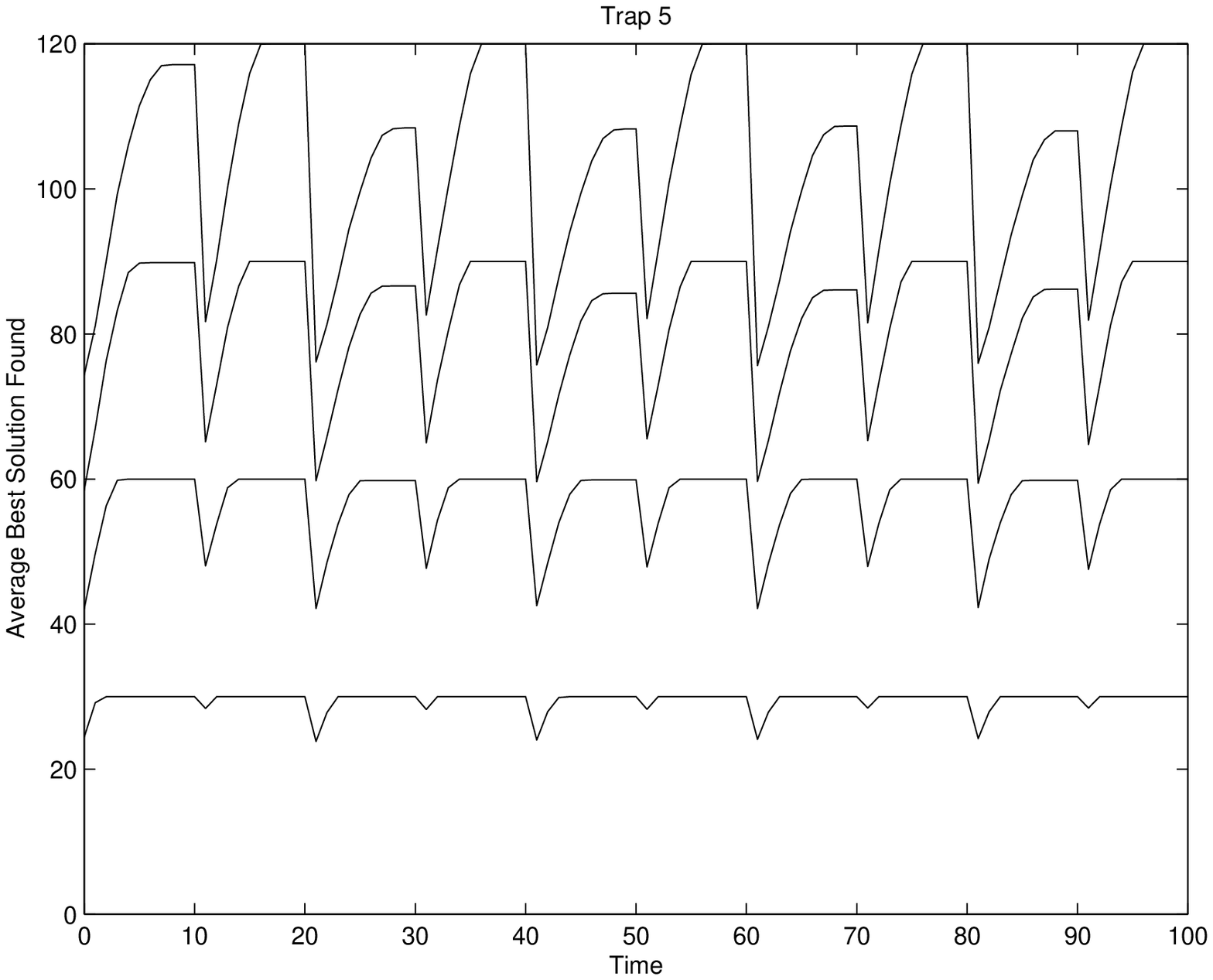,  width=2in, height=2in}
 \caption{Traps using dcGA(1) with last model (left) Cycle 5 (Right) Cycle 10. (Top) trap--3 (Middle) Trap--4 (Bottom) Trap--5. In each graph, the four curves correspond to 5, 10, 15, and 20 building blocks ordered from bottom up.}\label{res1}
\end{center}
\end{figure}

\begin{figure} [h!]
\begin{center}
 \epsfig{figure=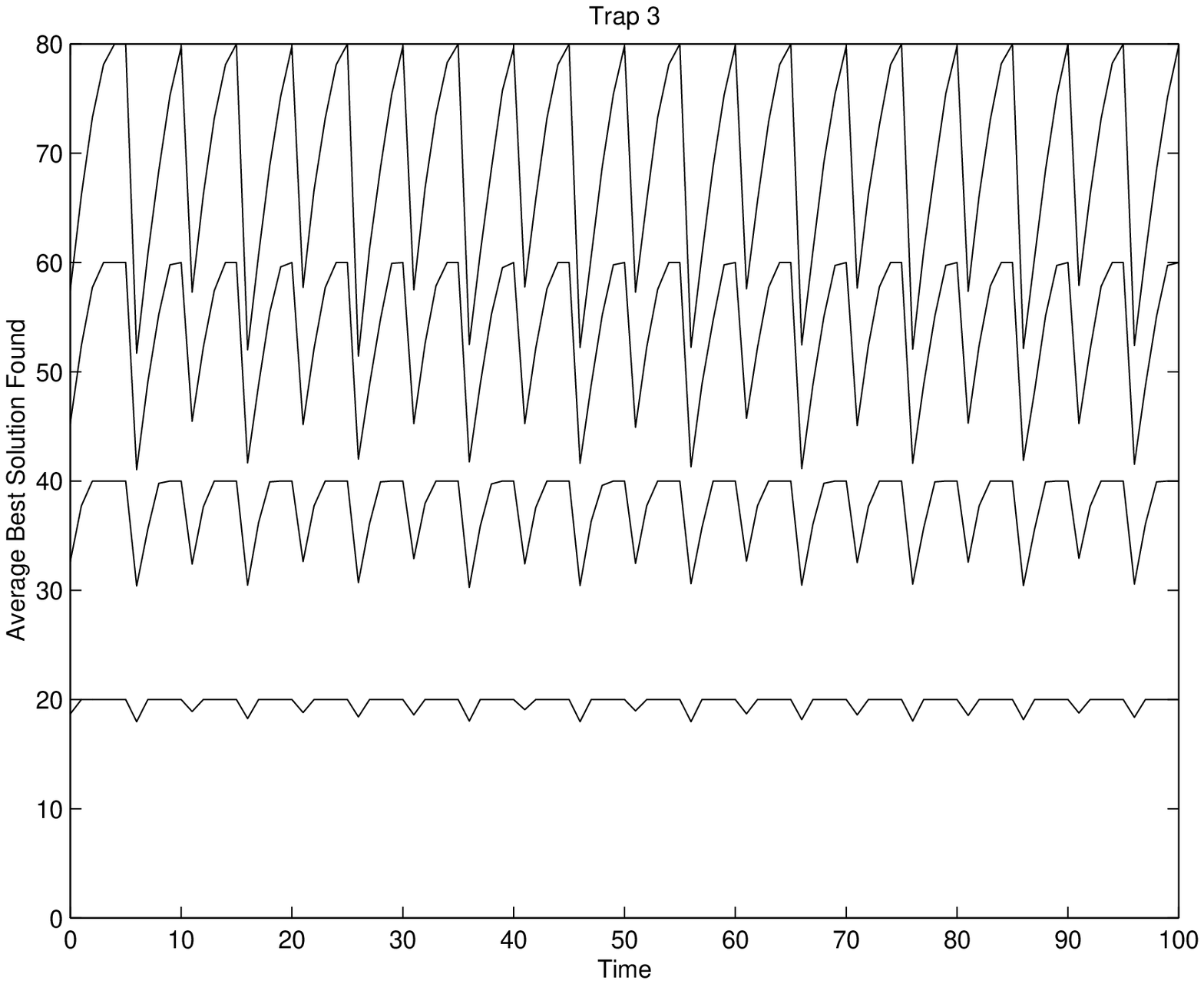,  width=2in, height=2in}
 \epsfig{figure=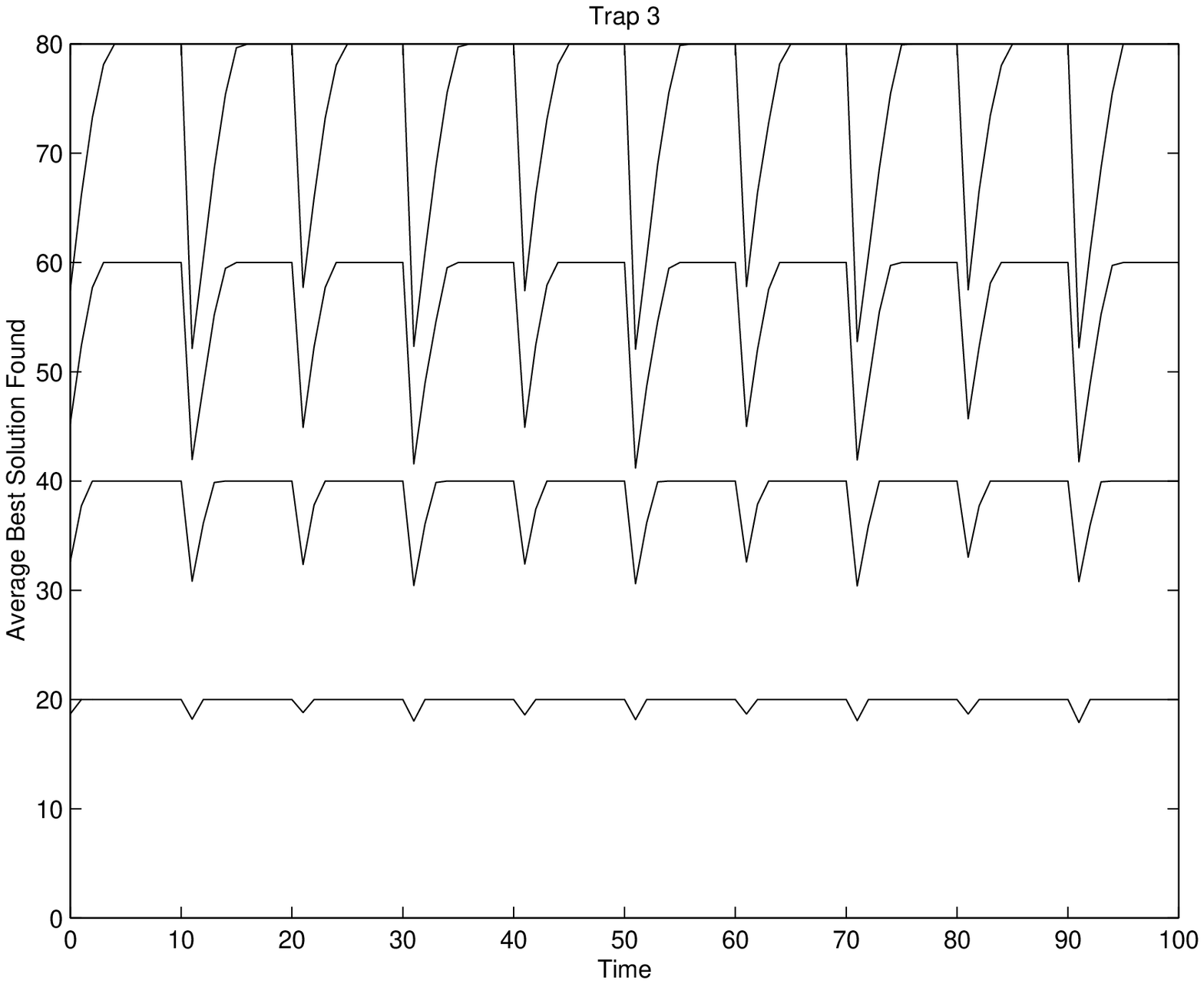,  width=2in, height=2in}

 \epsfig{figure=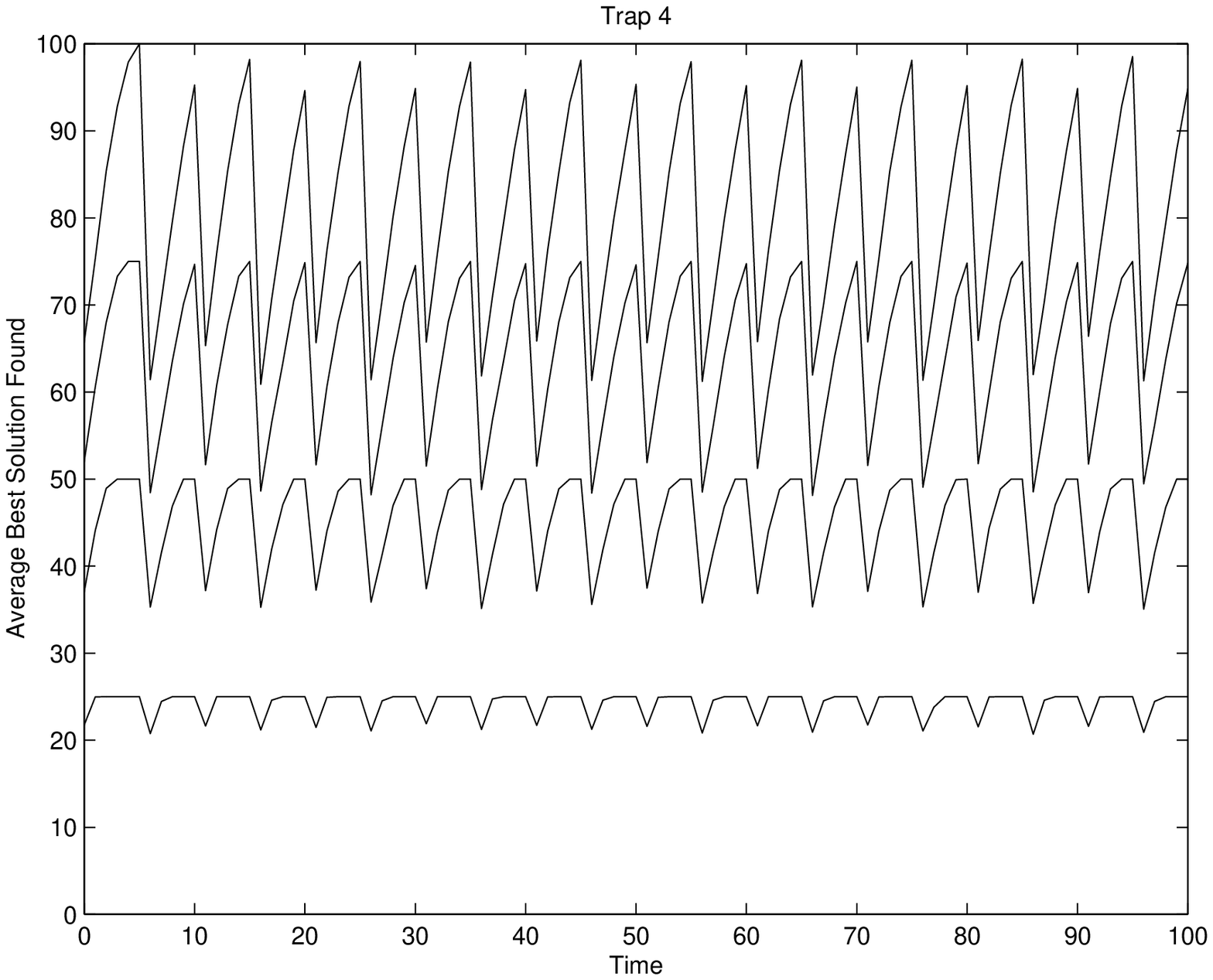,  width=2in, height=2in}
 \epsfig{figure=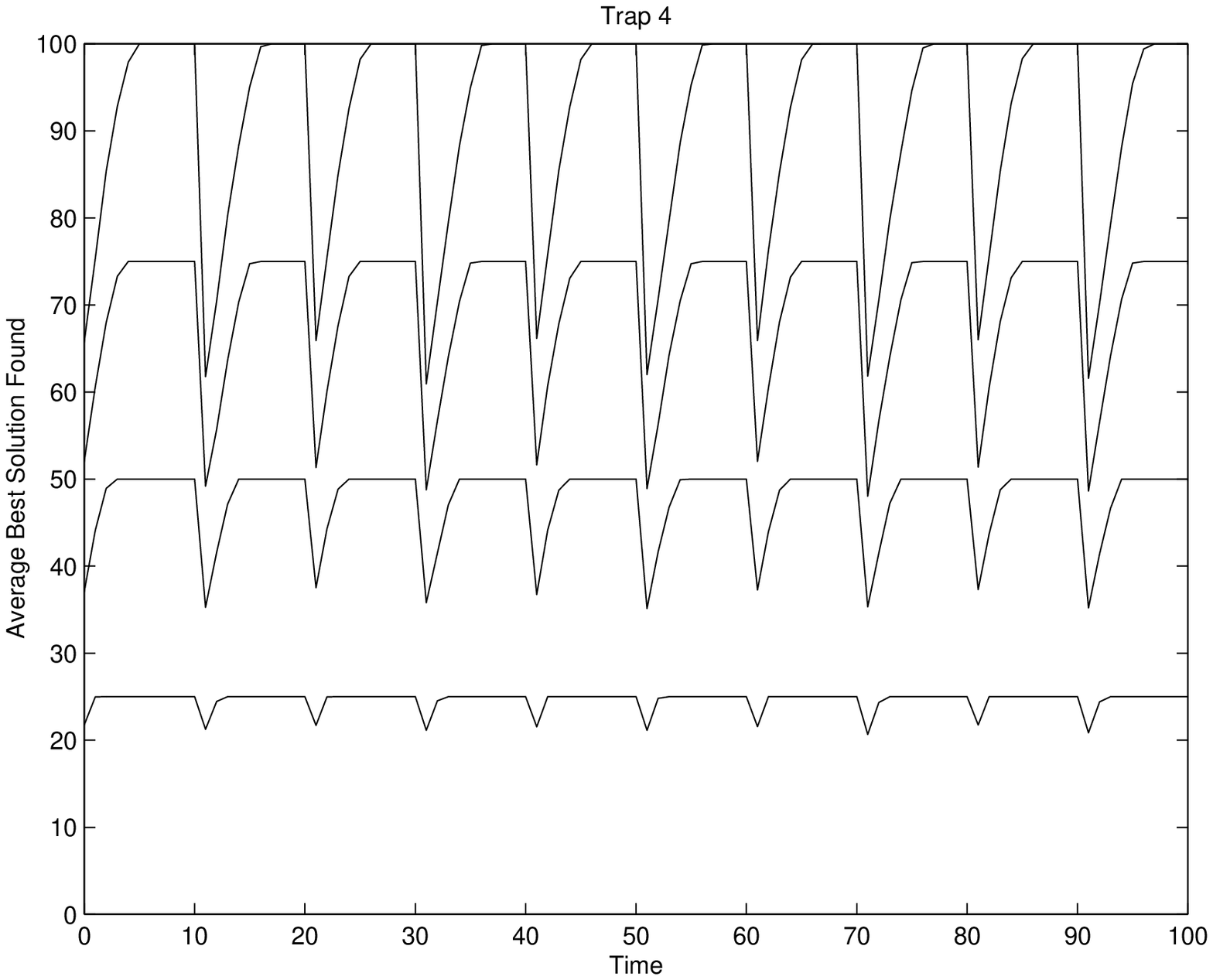,  width=2in, height=2in}

 \epsfig{figure=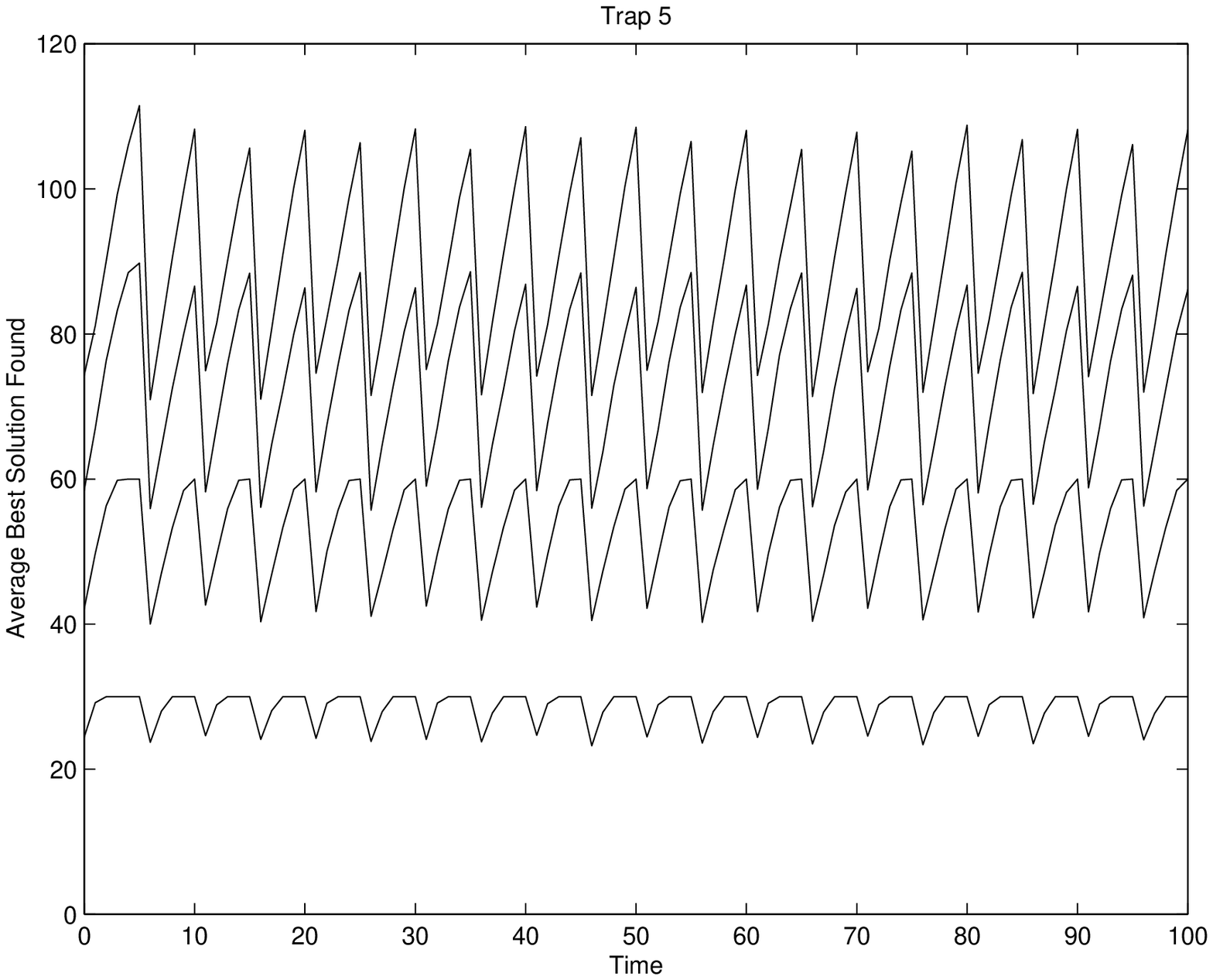,  width=2in, height=2in}
 \epsfig{figure=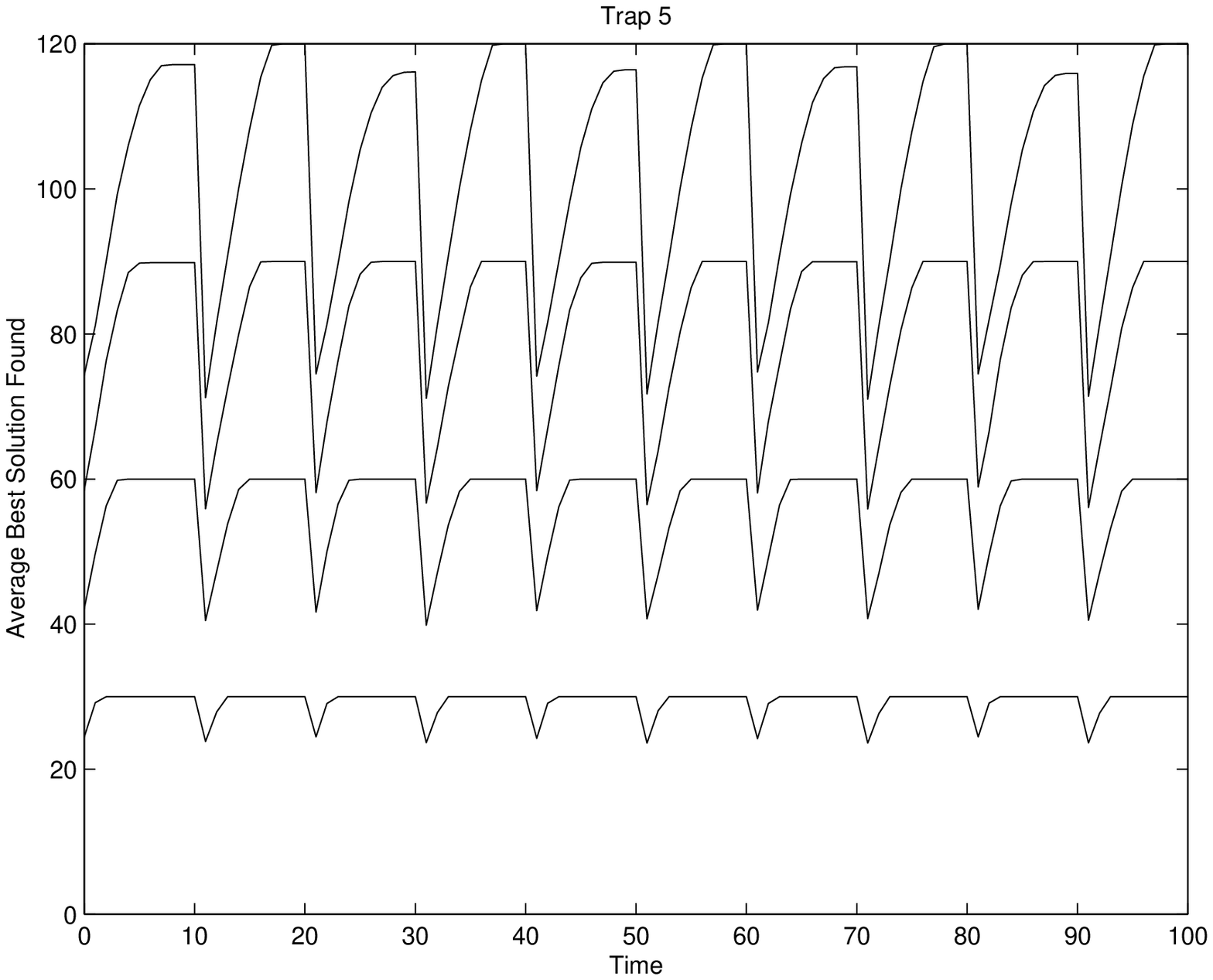,  width=2in, height=2in}
 \caption{Traps using dcGA(2) without last model (left) Cycle 5 (Right) Cycle 10. (Top) trap--3 (Middle) Trap--4 (Bottom) Trap--5. In each graph, the four curves correspond to 5, 10, 15, and 20 building blocks ordered from bottom up.}\label{res2}
\end{center}
\end{figure}

\begin{figure} [h!]
\begin{center}
 \epsfig{figure=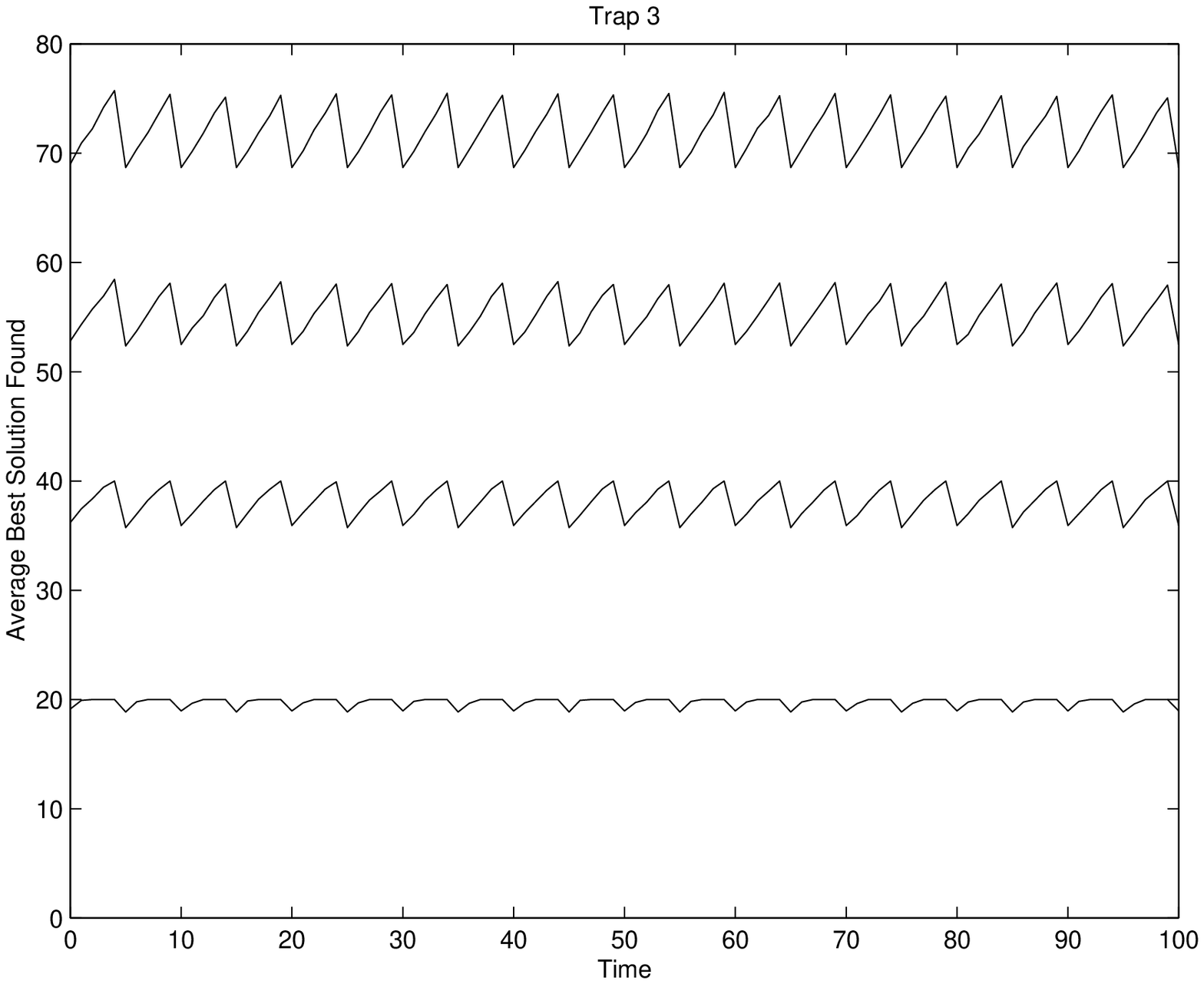,  width=2in, height=2in}
 \epsfig{figure=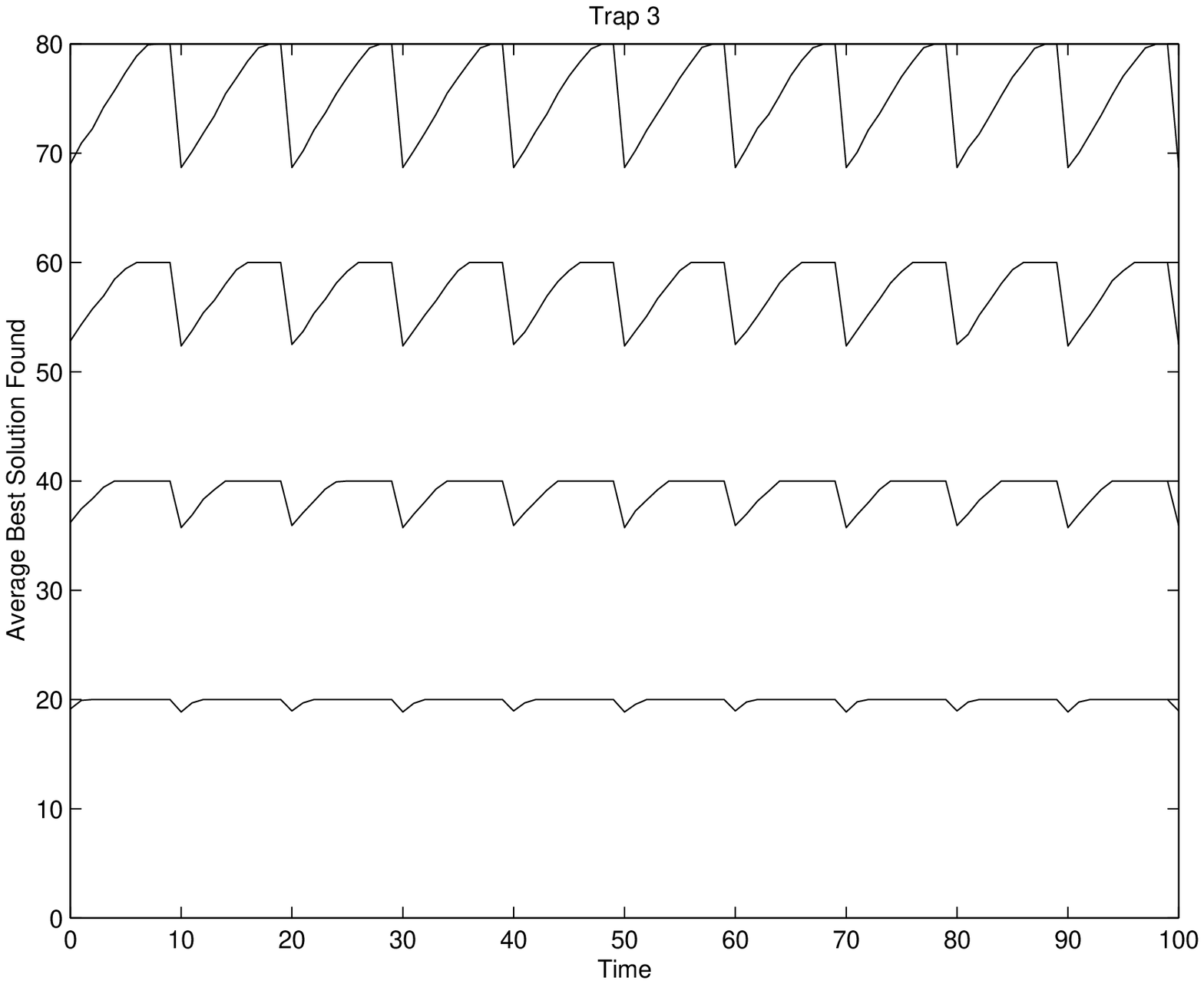,  width=2in, height=2in}

 \epsfig{figure=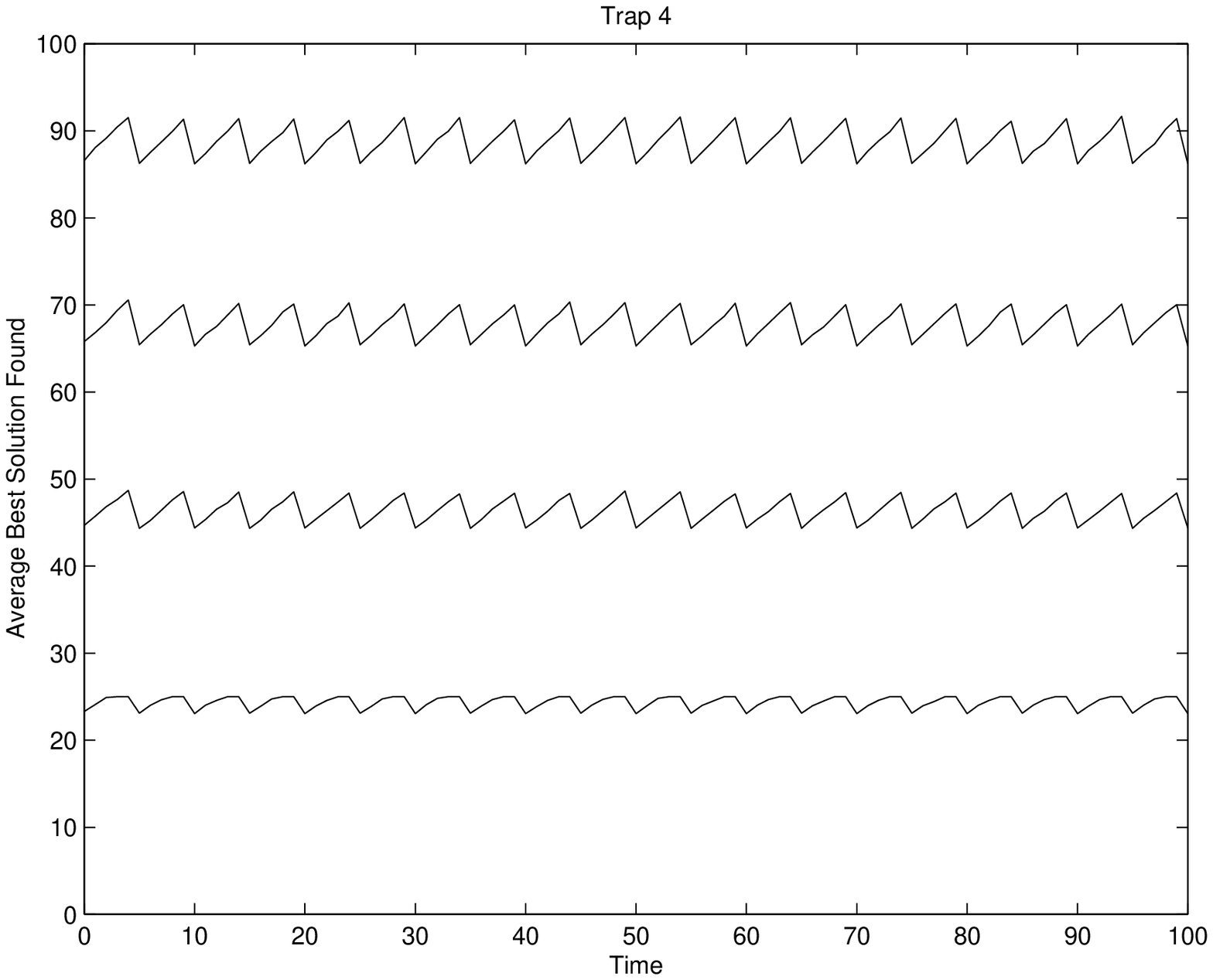,  width=2in, height=2in}
 \epsfig{figure=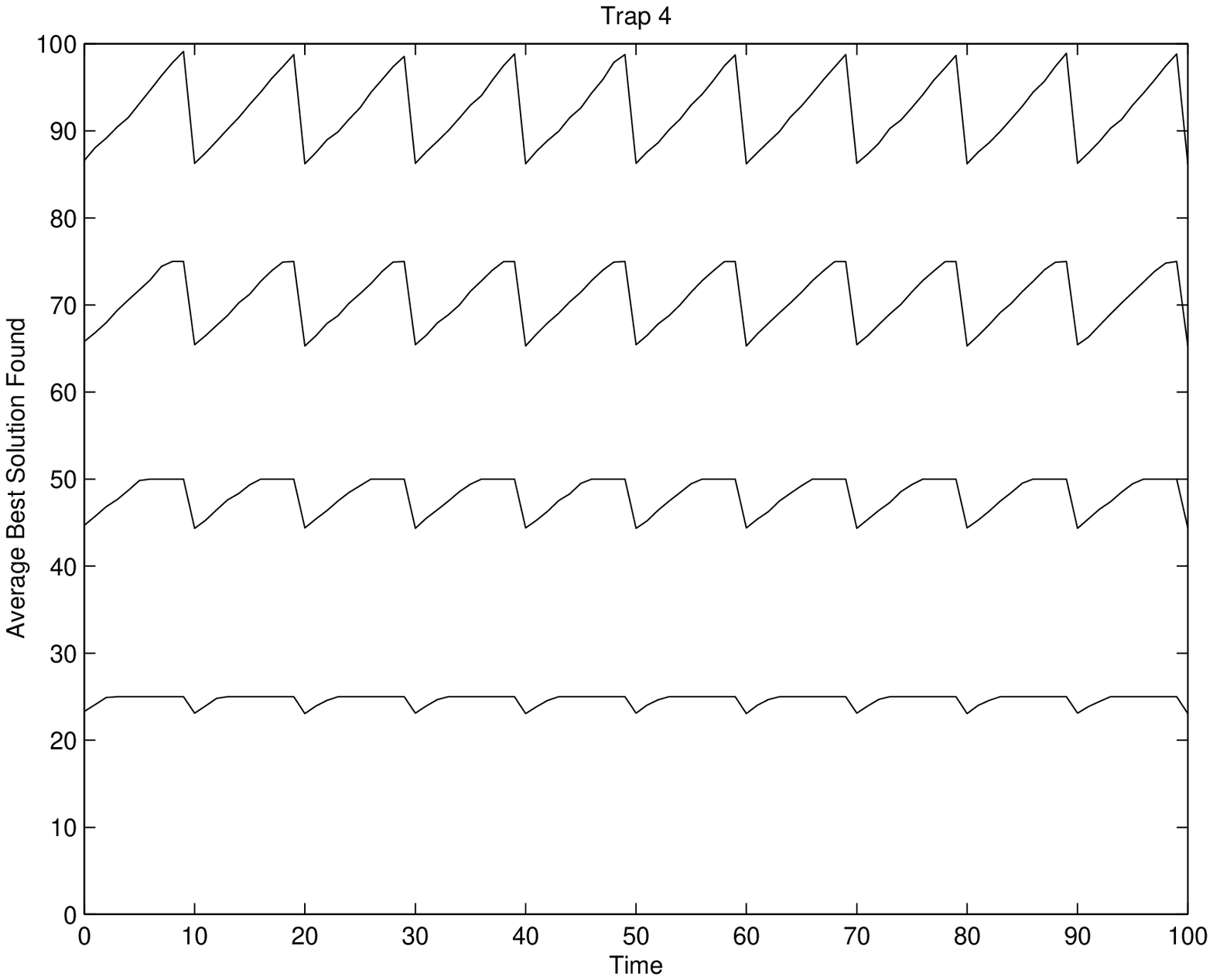,  width=2in, height=2in}

 \epsfig{figure=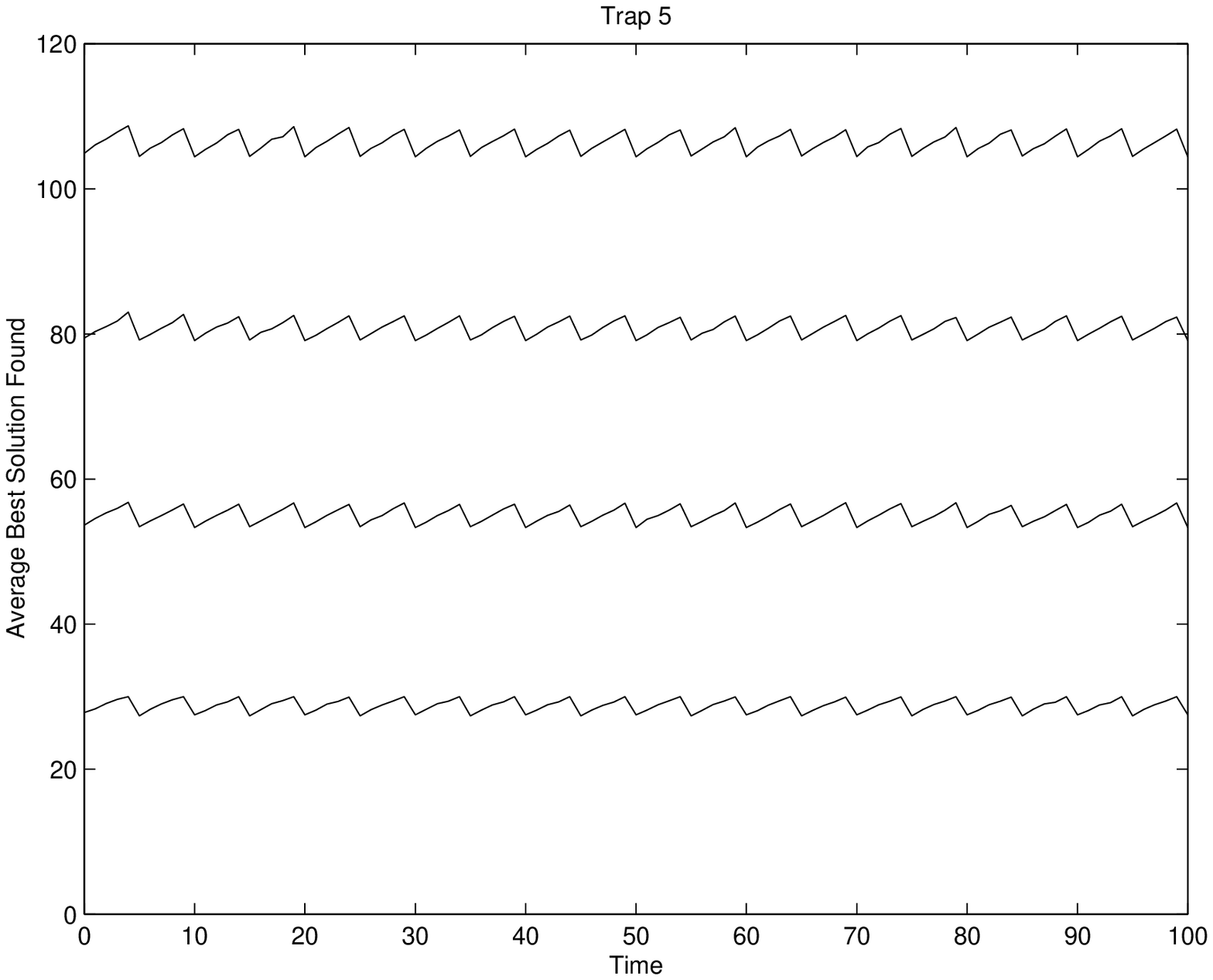,  width=2in, height=2in}
 \epsfig{figure=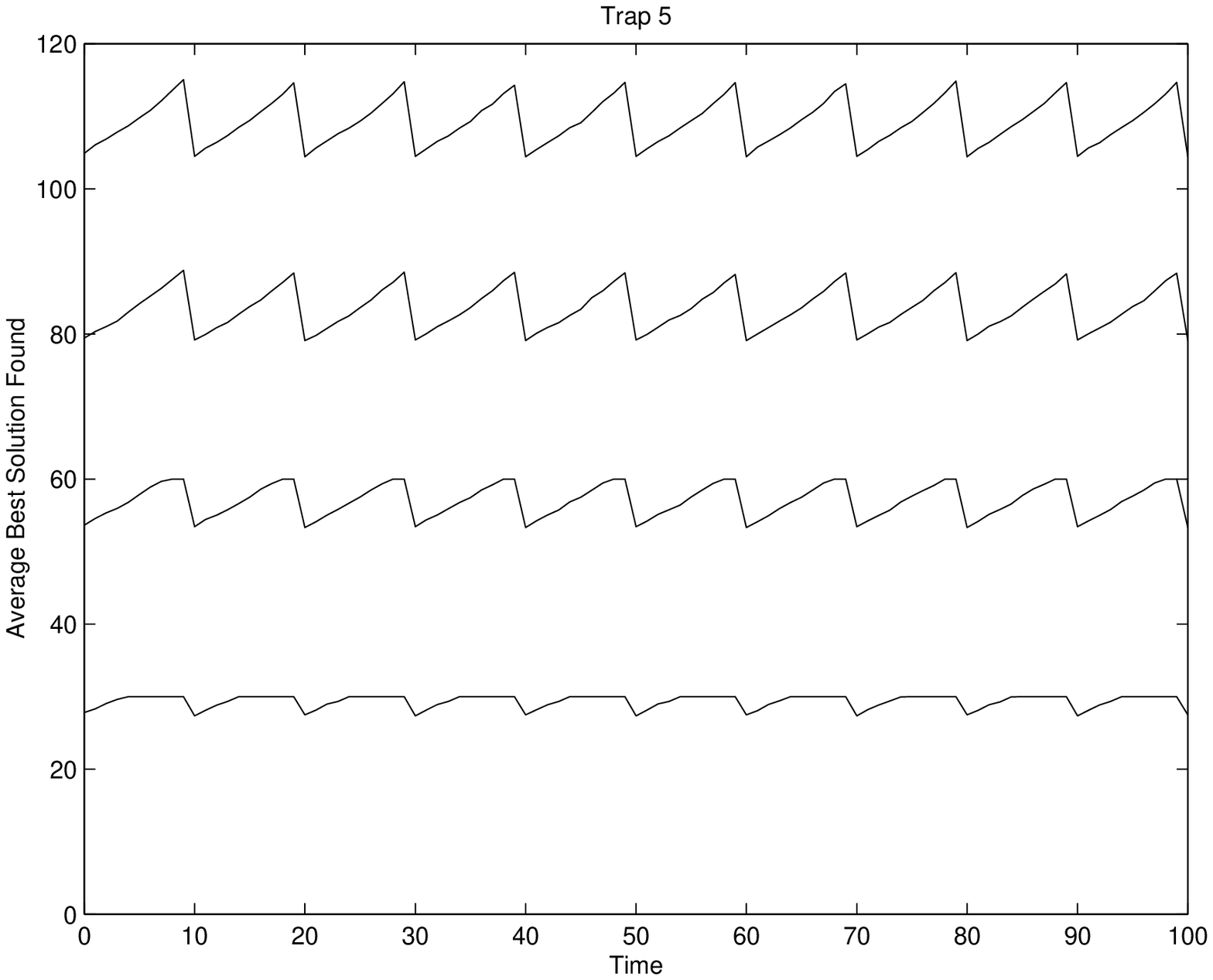,  width=2in, height=2in}
 \caption{Traps using uGA (left) Cycle 5 (Right) Cycle 10. (Top) trap--3 (Middle) Trap--4 (Bottom) Trap--5. In each graph, the four curves correspond to 5, 10, 15, and 20 building blocks ordered from bottom up.}\label{res3}
\end{center}
\end{figure}

Figures~\ref{res1},~\ref{res2}, and~\ref{res3} present the
performance of dcGA(1), dcGA(2) and uGA respectively. Starting
with the performance of dcGA(1) as being depicted in
Figure~\ref{res1}, we can see that the algorithm consistently
responds quickly to changes in the environment with trap--3
regardless of the number of building blocks, and cycle length.
However, we can see that the response rate with trap--4 is less as
indicated with the drop in performance with cycle length 5 and the
good performance with the longer cycle length of 10. From the
figure, it can be seen that the higher the order of the trap, the
slower the method is able to respond to a change in the
environment. It can also be seen that the larger the number of
copies of building blocks in the chromosome, the slower the
response to environmental changes. The slowest response rate was
encountered with trap--5 and 20 building blocks. These finding are
logical as the level of hardness in the problem increases as the
linkage and problem size increases. That is, the harder it is to
separate the variables, the more difficult it is to learn the
decomposition. This way, we can use the order of a trap and the
problem size to quantify how hard a dynamic optimization problem
is. \\

Similar patterns exist with the use of dcGA(2) as being depicted
in Figure~\ref{res2}. One can notice that the drop in performance
is less with dcGA(1) than it is the case with dcGA(2). Also, by
looking at trap--5 with cycle length 5, one can notice that the
performance of dcGA(2) is worse than the corresponding case using
dcGA(1). This is expected as the response rate would be higher
when using dcGA(1) as compared to dcGA(2); thanks to the bias in
the initial population with the last linkage model found. However,
by comparing trap--5 with cycle length of 10 using dcGA(2) against
the corresponding performance using dcGA(1), one can see that the
performance of dcGA(2) is consistently better than the
corresponding performance of dcGA(1). An explanation of this
result will be presented in the following subsection. \\

Comparing the previous results to the uGA results which are
depicted in Figure~\ref{res2}, unexpectedly one can see that uGA
is very competitive to the linkage learning model. After careful
examination of the performance of uGA, we identified that the key
reason behind the success of uGA is due to simple luck. As the two
attractors in the problem exist when all solutions are 0's and
1'0, if uGA converges to the wrong attractor in one cycle, the
wrong attractor becomes the right attractor in the following cycle
and as it converges back to the wrong attractor, the environment
changes again and switches the wrong attractor to become the
preferred attractor. In other words, the environment changes in a
manner that is beneficial for the bad performance of uGA. To
verify our analysis, we conducted a second experiment as being
explained in the following subsection. \\

\subsection{Experiment 2}

\begin{figure} [h!]
\begin{center}
 \epsfig{figure=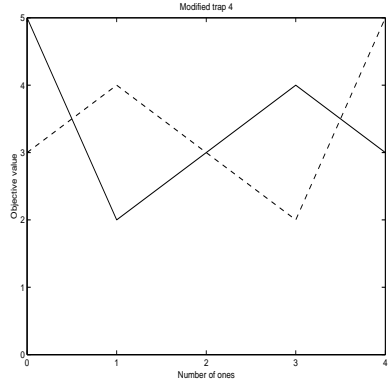, width=2in, height=2in}
 \caption{The modified trap function 4 in a changing environment.}\label{ftrap2}
\end{center}
\end{figure}

In the second type of experiments, we modified the trap function
of order 4 to break the symmetry in the attractors. The new
function is visualized in Figure~\ref{ftrap2}. At time 0 and in
even cycles, the optimal solution is when all variables are set to
0's and the second attractor is when the sum of 1's is equal to 3.
When the environment changes during the odd cycles, the new
solution is optimal when all variables set to 1's with a new
deceptive attractor when the sum of 1's is 1 or alternatively the
number of 0's is 3. This setup guarantees that the trap is not
symmetric with regards to its attractors. Some researchers
suggested that a simple use of an Xor operator with trap functions
would solve the problem easily because once the GA method
converges to the wrong attractor, a simple Xor operator would take
it to the right attractor. In our design in Figure~\ref{ftrap2},
breaking the symmetry in the trap would also counterpart the
possible trick of using an Xor operator.\\

\begin{figure} [h!]
\begin{center}
 \epsfig{figure=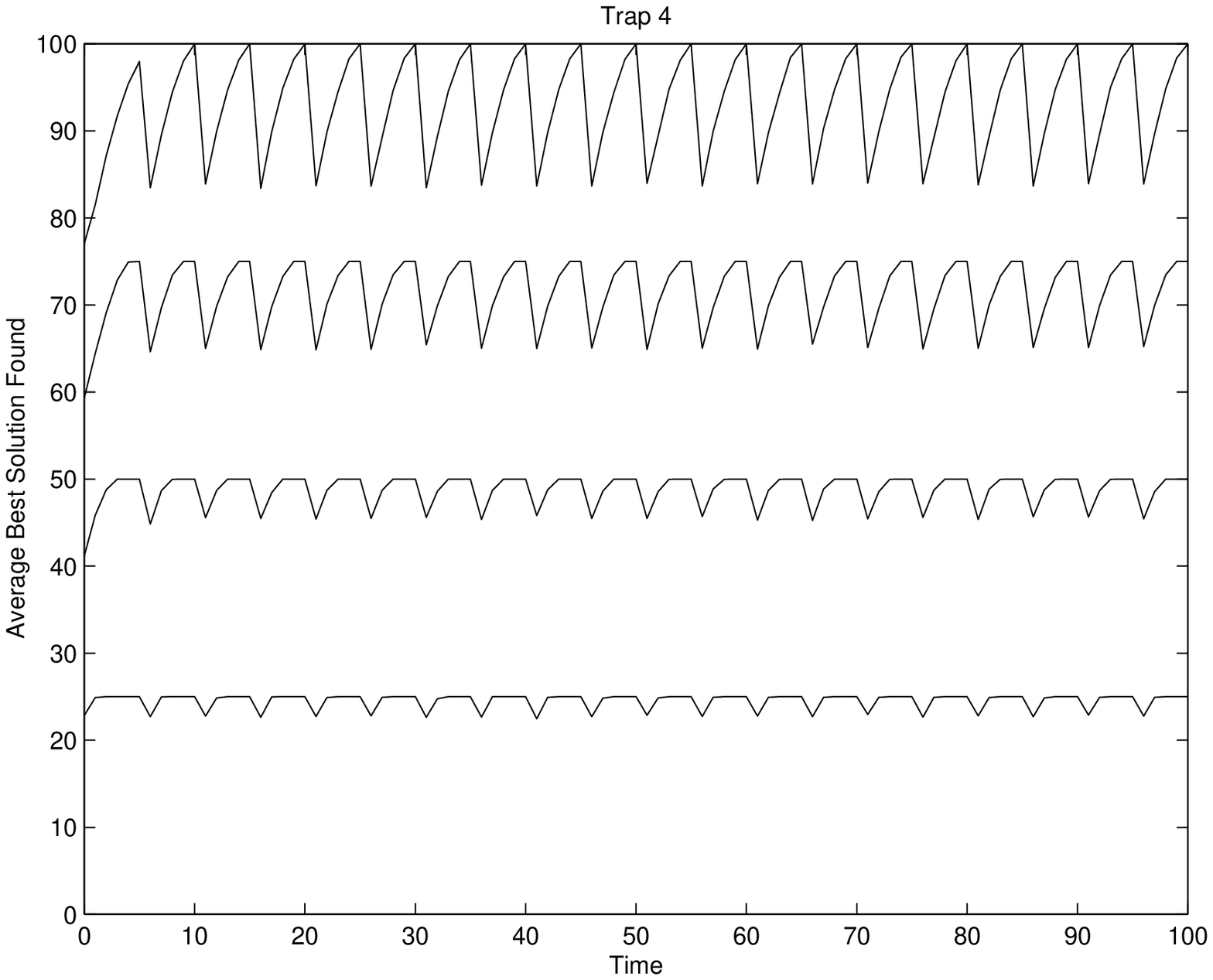,  width=2in, height=2in}
 \epsfig{figure=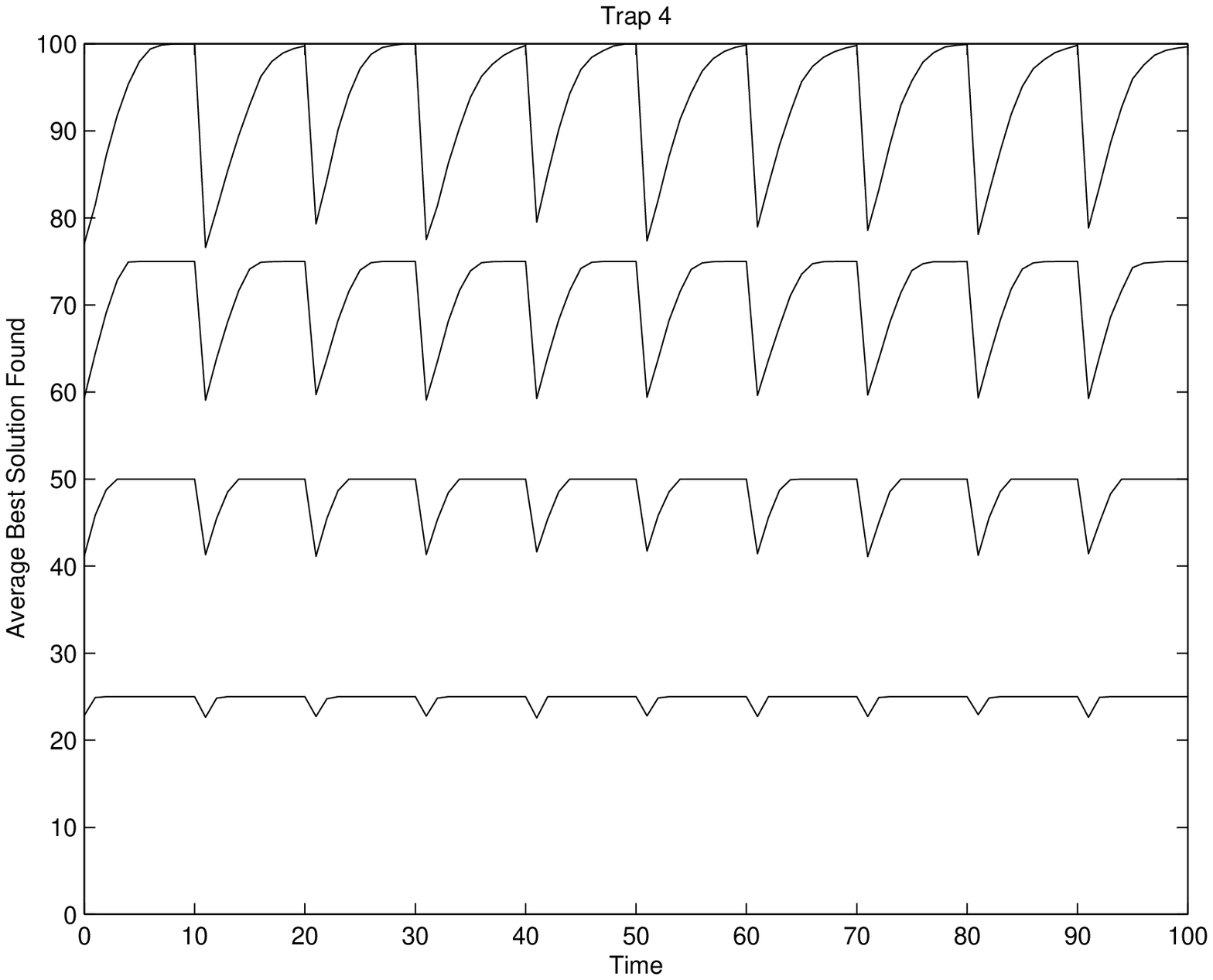,  width=2in, height=2in}

 \epsfig{figure=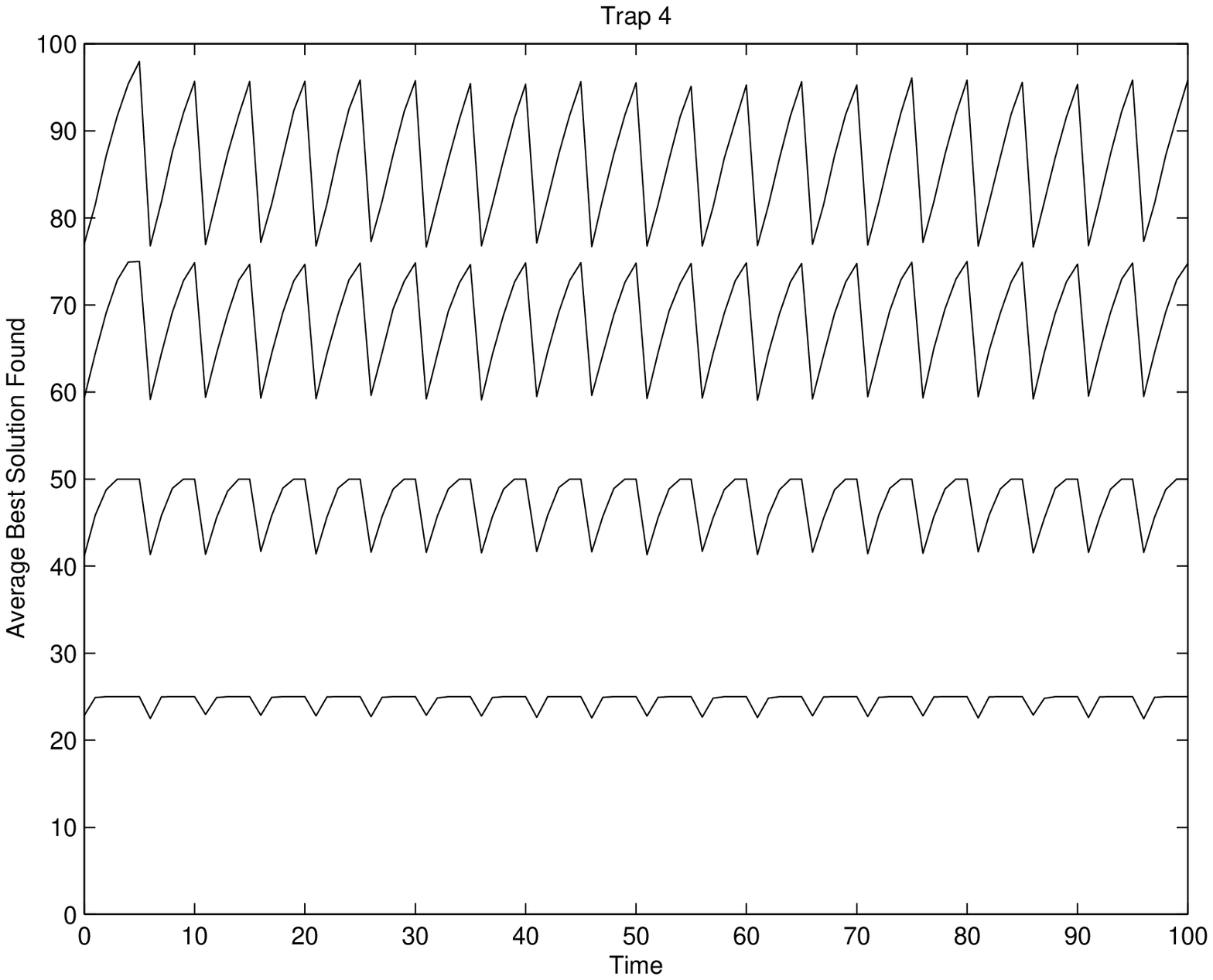,  width=2in, height=2in}
 \epsfig{figure=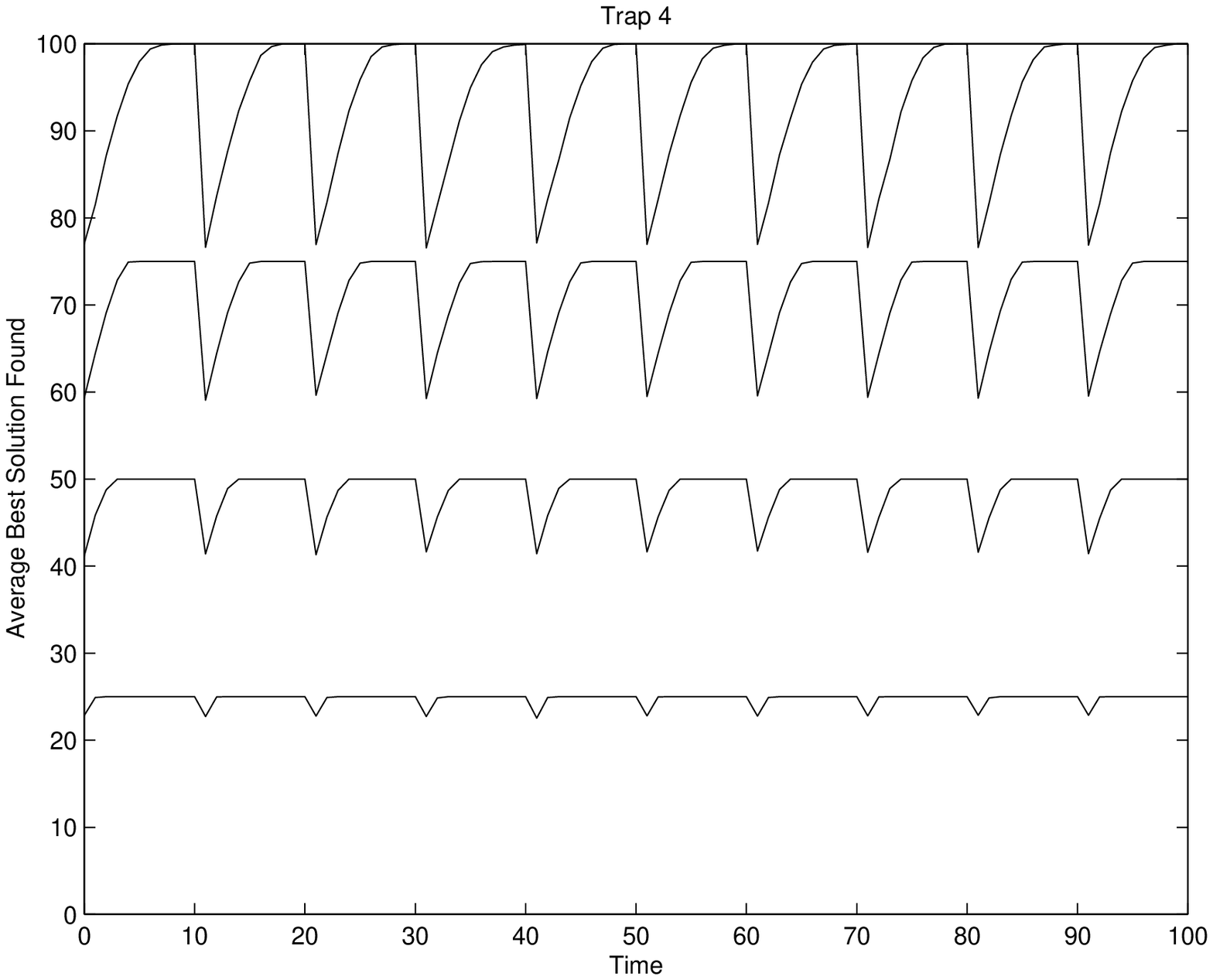,  width=2in, height=2in}

 \epsfig{figure=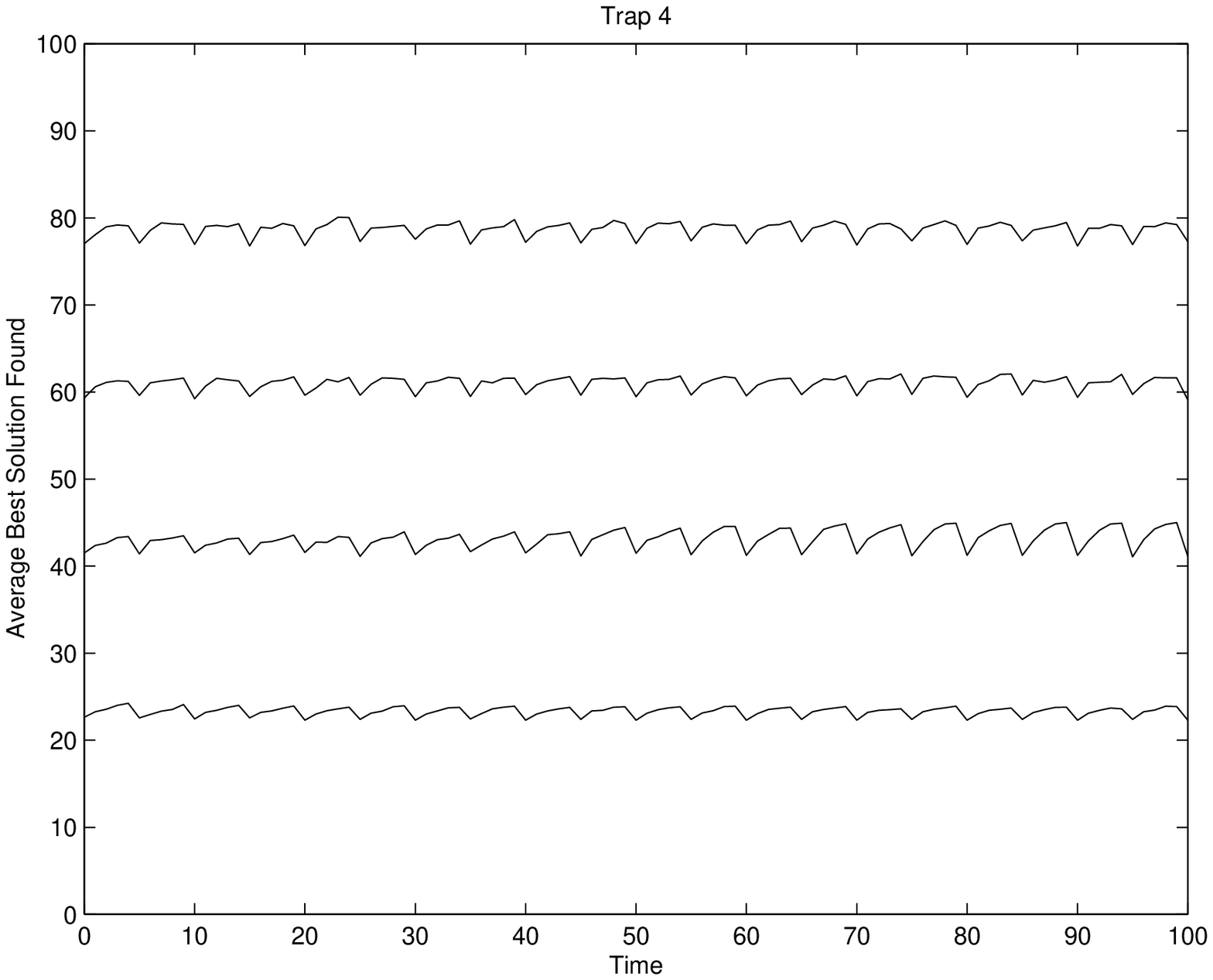,  width=2in, height=2in}
 \epsfig{figure=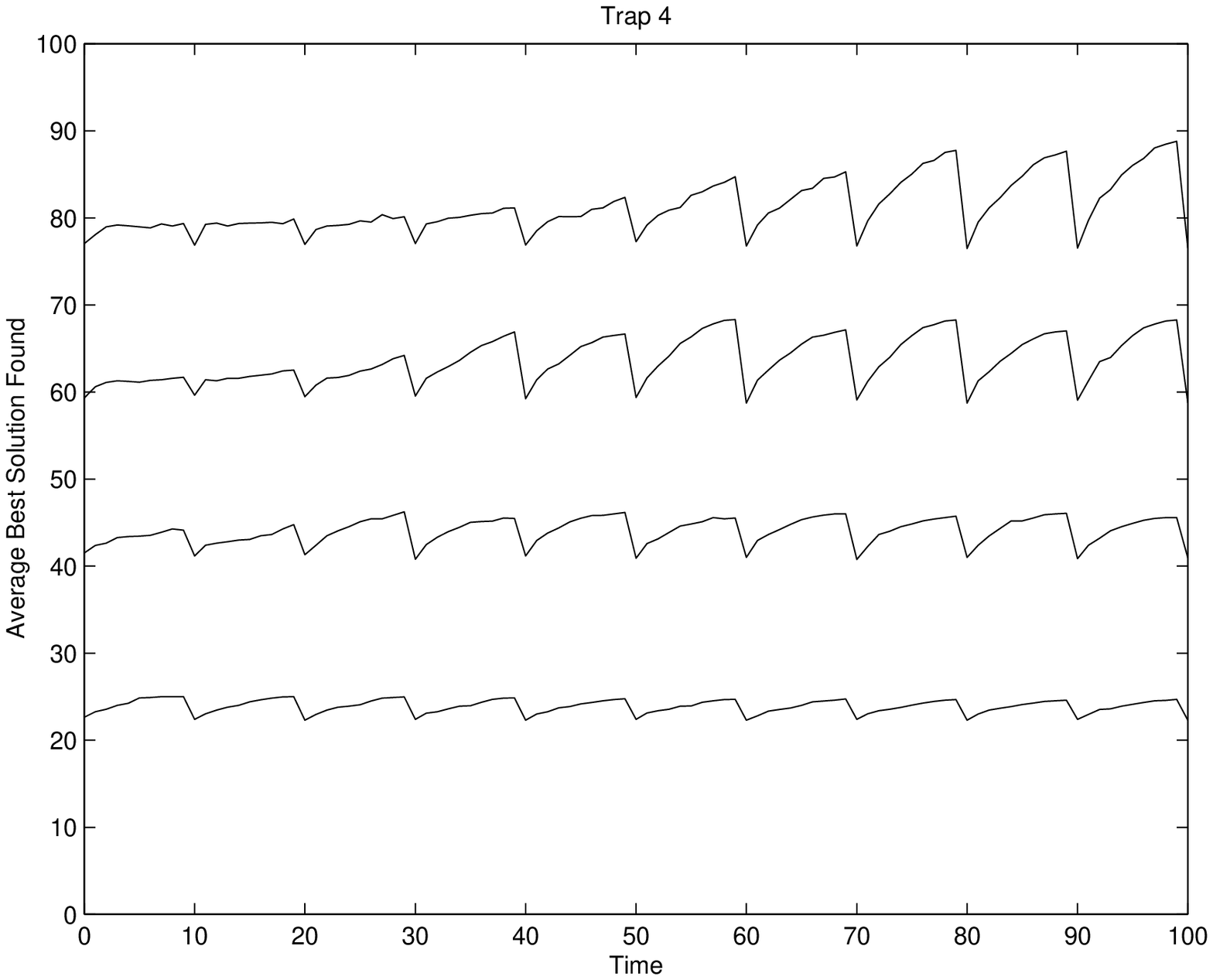,  width=2in, height=2in}
 \caption{Modified Trap 4 (left) Cycle 5 (Right)
 Cycle 10, (top) dcGA(1), (middle) dcGA(2), (Bottom) uGA. In each graph, the four curves
correspond to 5, 10, 15, and 20 building blocks ordered from
bottom up.}\label{res4}
\end{center}
\end{figure}

Figure~\ref{res4} depicts the behavior of the three methods using
the modified trap--4 function. As expected, the uGA method clearly
shows the worst behavior among the three methods. It is clear that
it is unable to respond to the changes neither it is able to even
get to the deceptive attractor in some cases. This behavior
confirms our analysis in the previous section. When looking at
dcGA(1) and dcGA(2), however, we can see that dcGA(1) is better
than dcGA(2). The dcGA(1) method is able to respond to the changes
in the environment quickly, accurately, and reliably all the time.
This result is somehow different as compared to the results
obtained from the previous section. The linkage has not changed
between the two setups, but the only change took place for the
attractors. This suggests that the cause of the somehow inferior
performance of dcGA(1) as compared to dcGA(2) is attributed to the
crossover operator or mixing strategy that it was slow in reaching
the two attractors with maximum hamming distance in the previous
experiments.\\

\subsection{Experiment 3}

\begin{figure} [h!]
\begin{center}
 \epsfig{figure=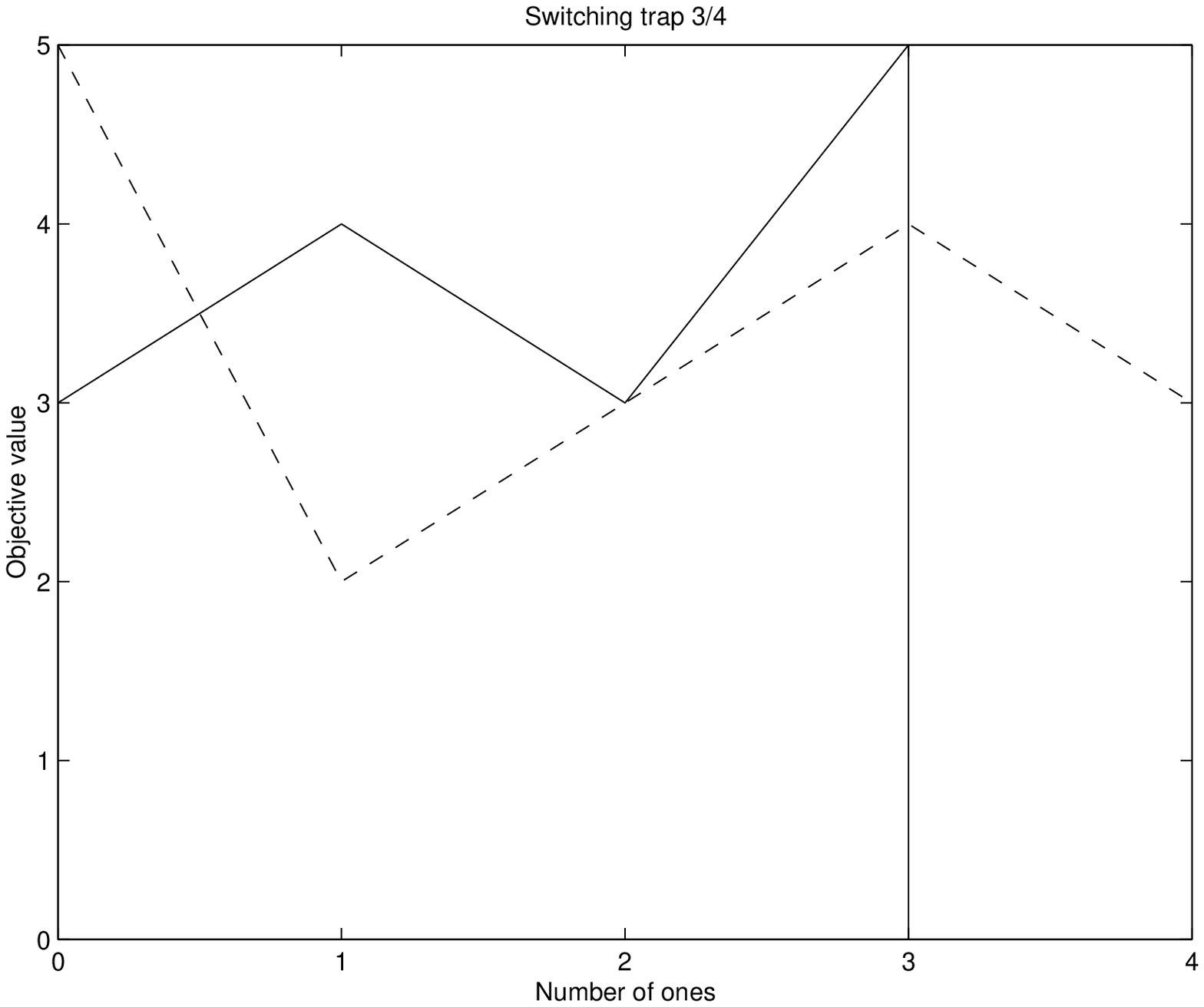, width=2in, height=2in}
 \caption{The switching trap function with k=3,4 in a changing environment.}\label{ftrap3}
\end{center}
\end{figure}

In this experiment, we subjected the environment under a severe
change from linkage point of view. Here, the linkage boundary
changes as well as the attractors. As being depicted in
Figure~\ref{ftrap3}, the environment is switching between trap--3
with all optima at 1's and trap--4 with all optima at 0's.
Moreover, in trap--3, a deceptive attractor exists when the number
of 1's is 1 while in trap--4, a deceptive attractor exists when
the number of 1's is 3. This setup is tricky in the sense that, if
a hill climber gets trapped at the deceptive attractor for
trap--4, the behavior will be good for trap--3. However, this
hill--climber won't escape this attractor when the environment
switches back to trap-4 since the solution will be surrounded with
solutions of lower qualities. This setup tests also whether any of
the methods is behaving similar to a hill--climber.\\

\begin{figure} [h!]
\begin{center}
 \epsfig{figure=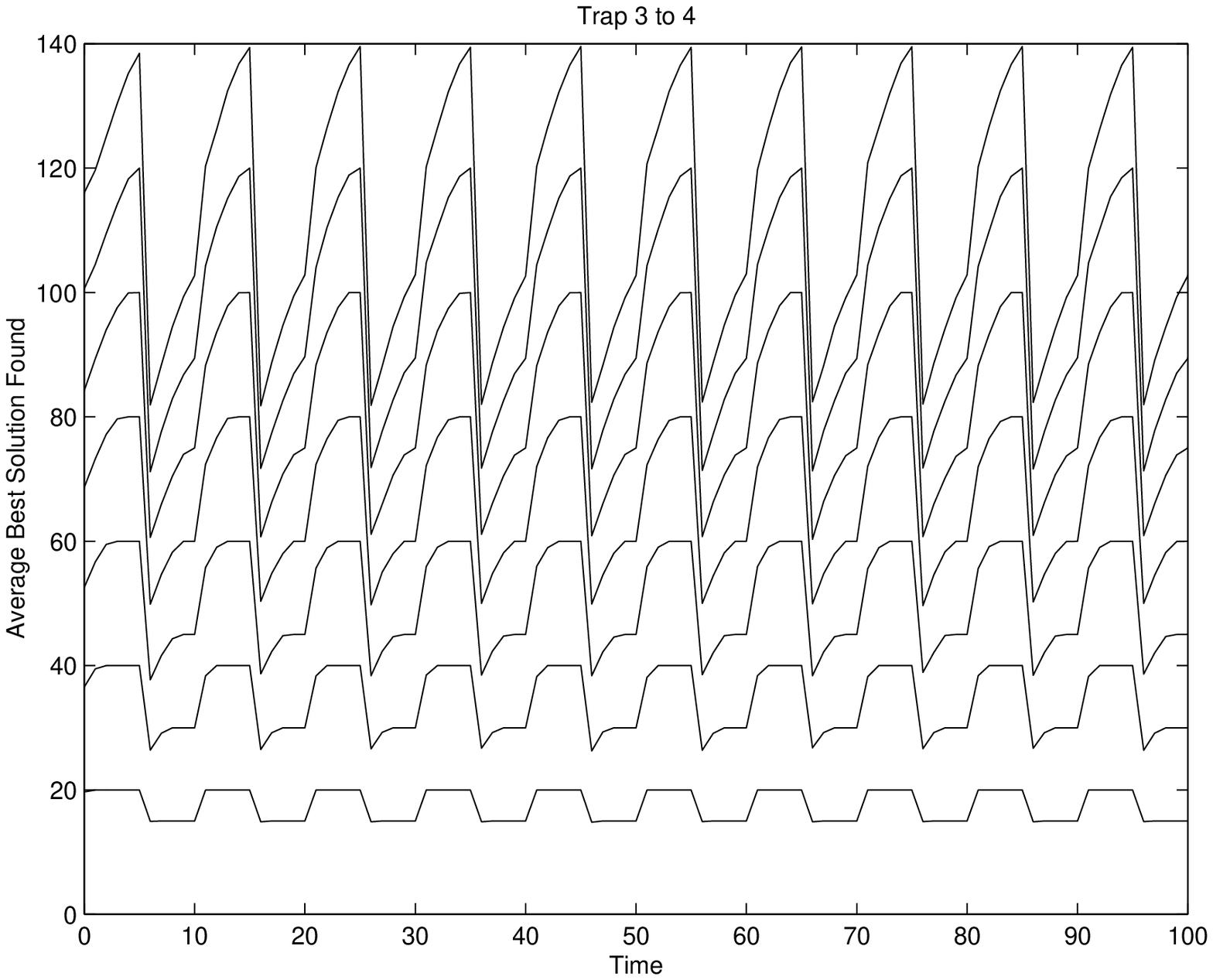,  width=2in, height=2in}
 \epsfig{figure=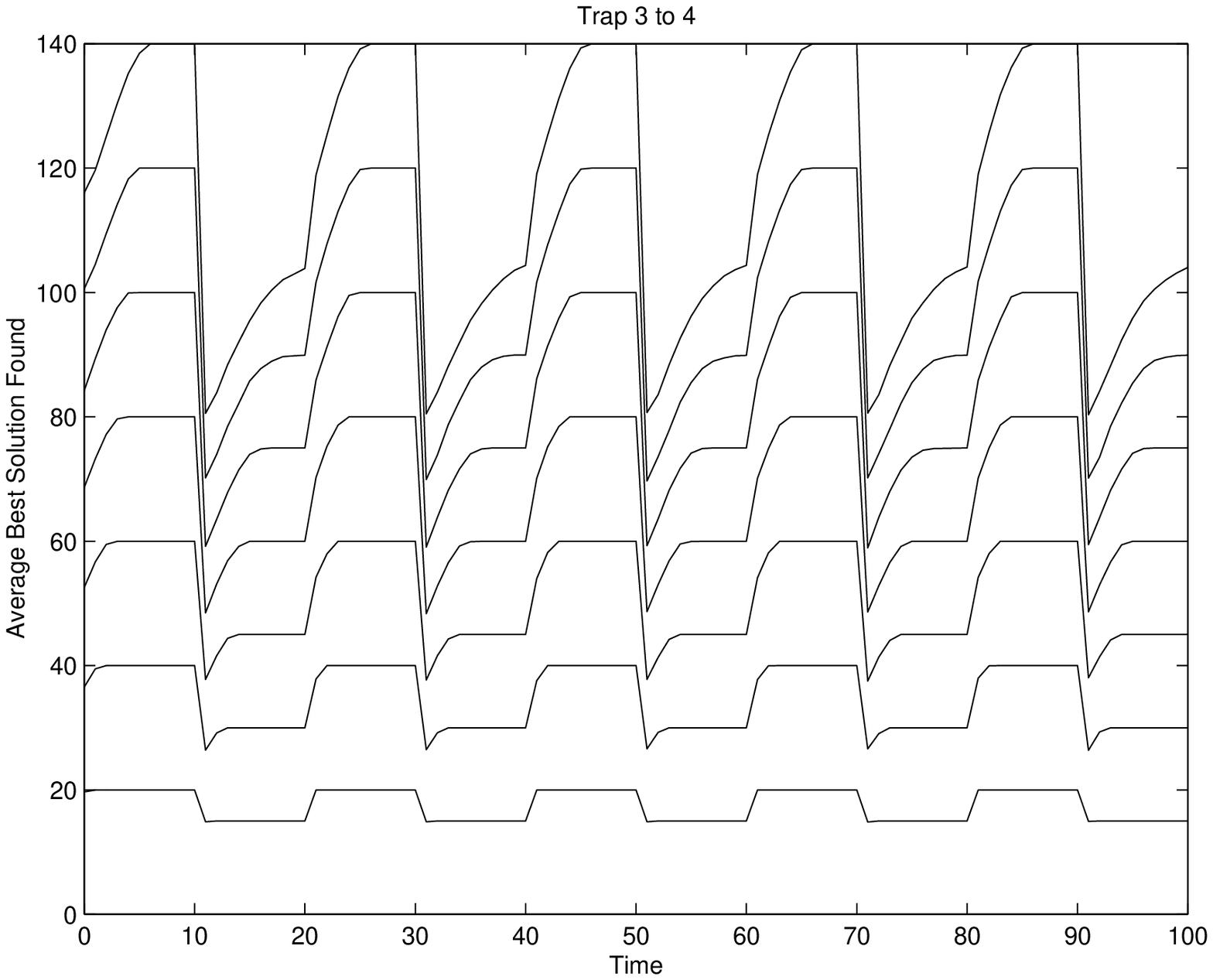,  width=2in, height=2in}

 \epsfig{figure=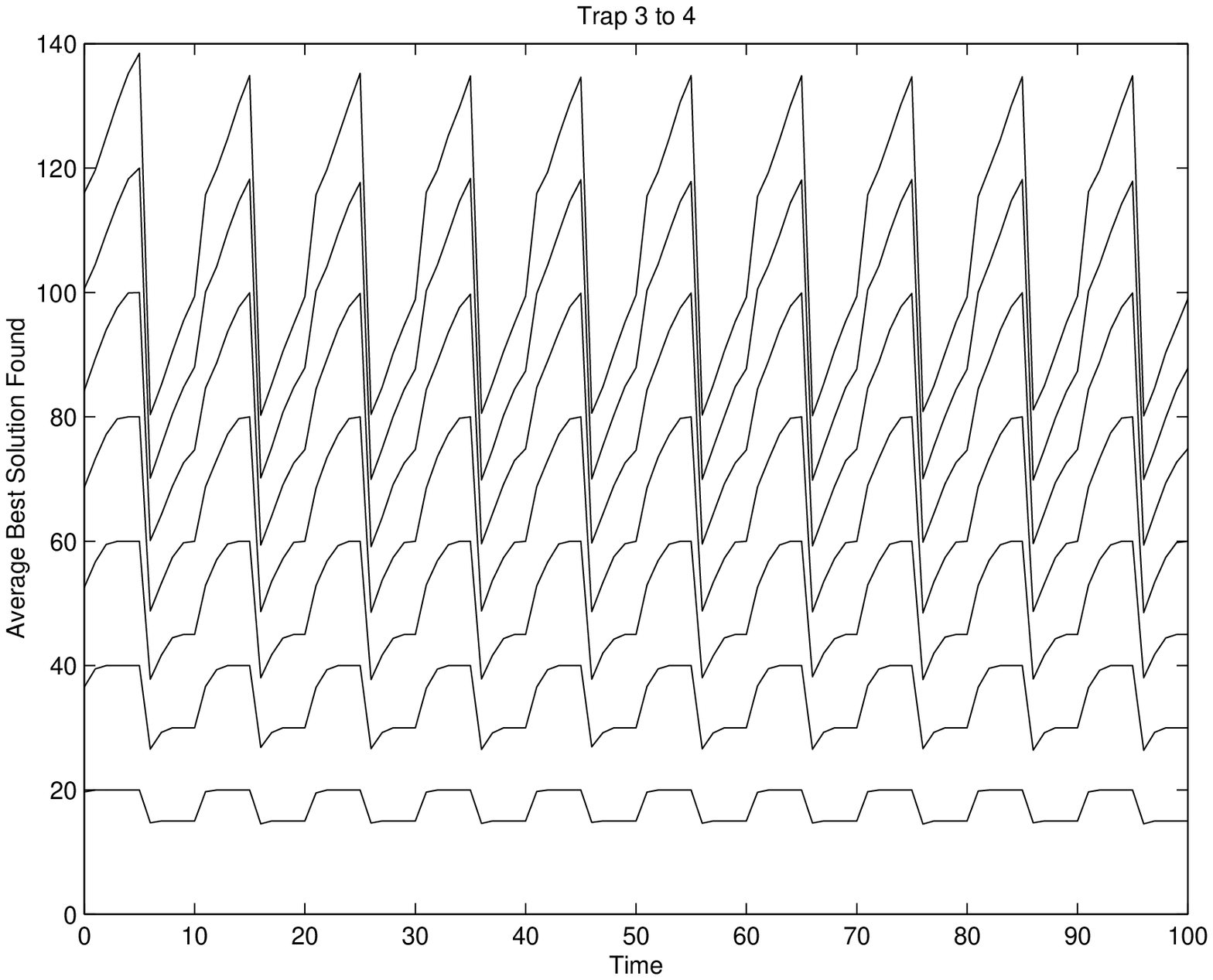,  width=2in, height=2in}
 \epsfig{figure=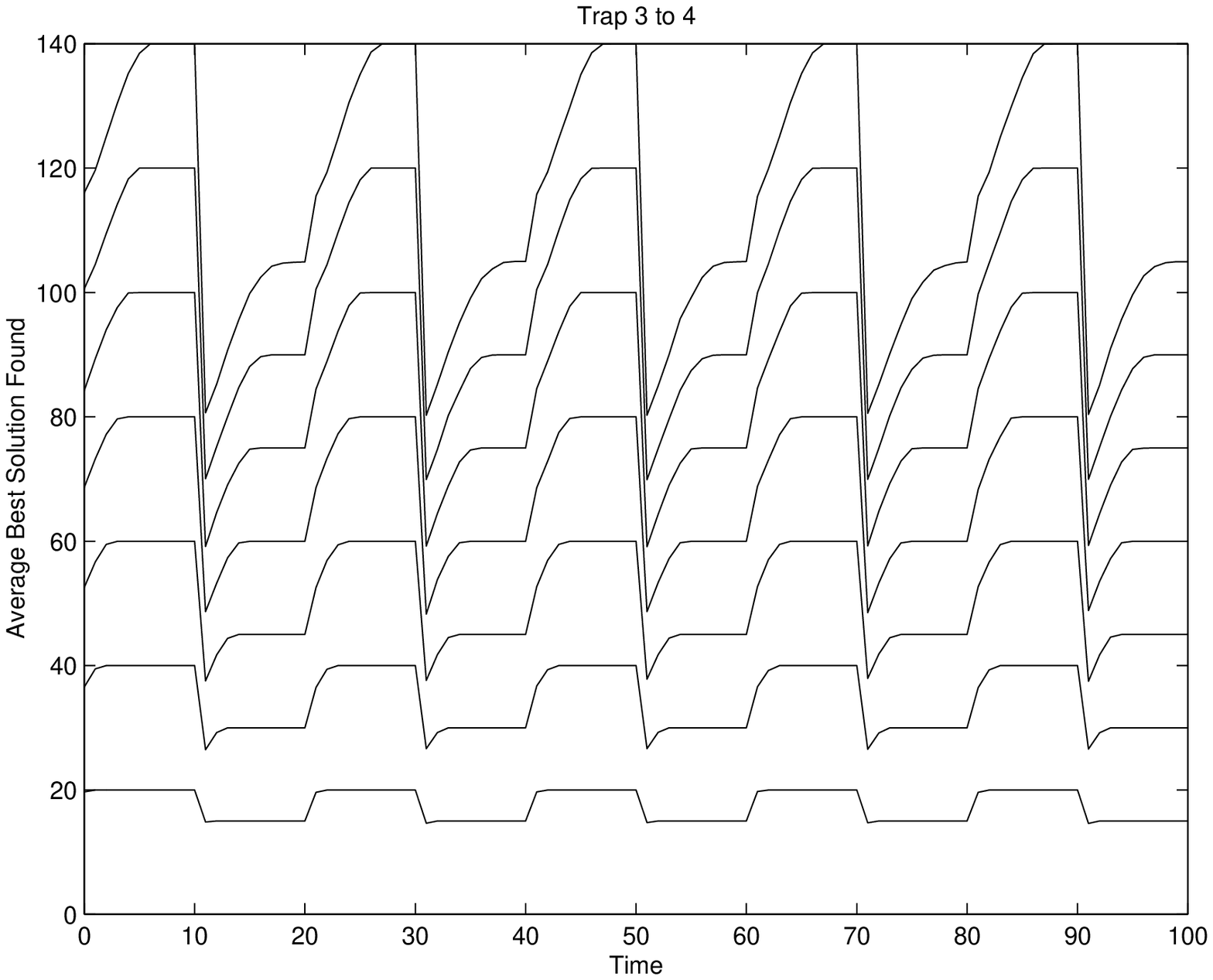,  width=2in, height=2in}

 \epsfig{figure=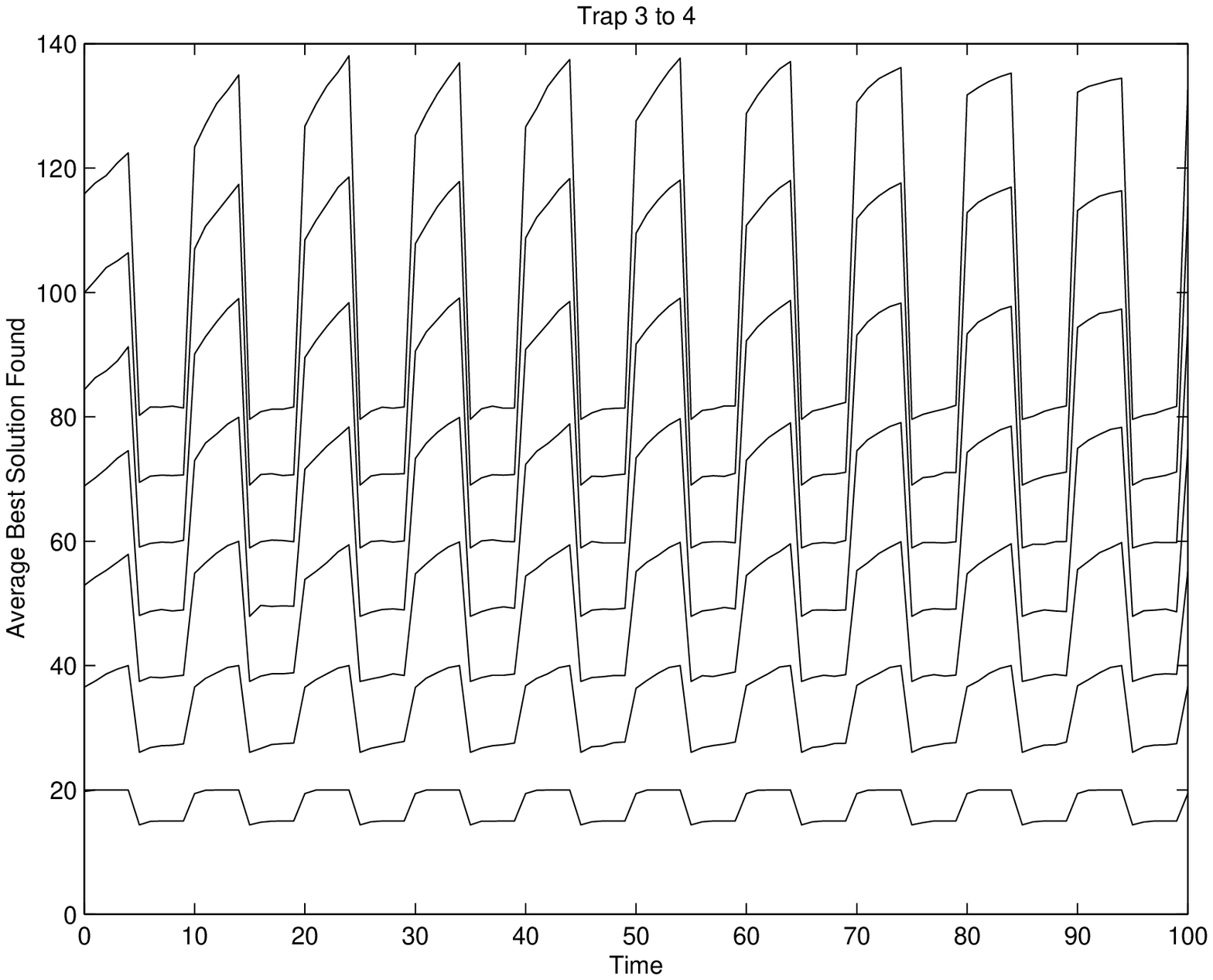,  width=2in, height=2in}
 \epsfig{figure=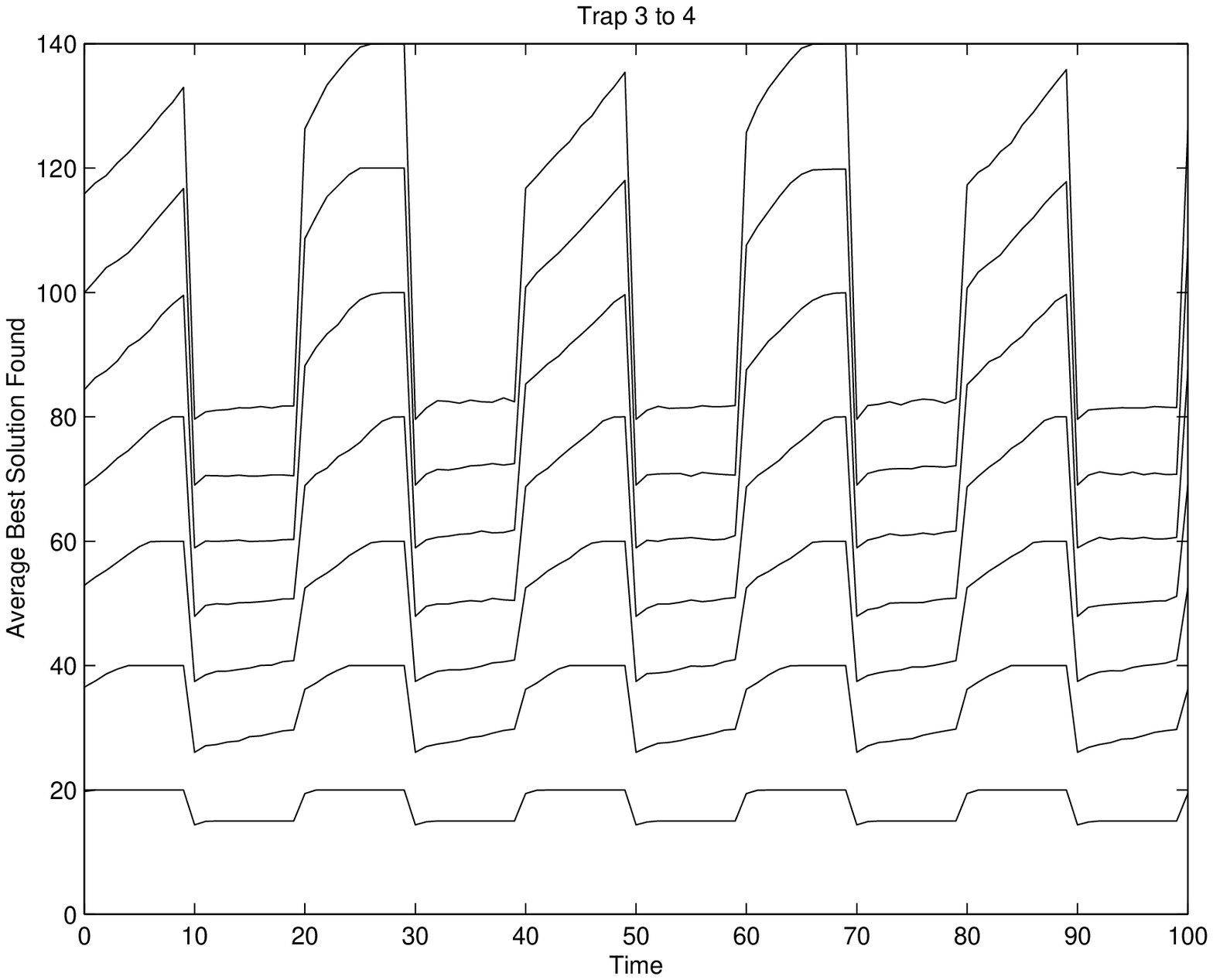,  width=2in, height=2in}
 \caption{Switching Trap 3--4 (left) Cycle 5 (Right)
 Cycle 10, (top) dcGA(1), (middle) dcGA(2), (Bottom) uGA. In each graph, the seven curves correspond to strings of length 12, 24, 36, 48, 60, 72, and 84 bits ordered from bottom up.}\label{res5}
\end{center}
\end{figure}

Figure~\ref{res5} shows the performance of dcGA(1), dcGA(2), and
uGA. We varied the string length between 12 and 84 in a step of 12
so that the string length is dividable by 3 and 4 (the order of
the trap). The following table lists the value of the optimal
solution for each string length with trap-3 and trap-4.\\

\begin{center}
\begin{tabular}{lcc}
\hline String Length & Trap--4 & Trap--3 \\ \hline
12  &   15  &   16  \\
24  &   30  &   32  \\
36  &   45  &   48  \\
48  &   60  &   64  \\
60  &   75  &   80  \\
72  &   90  &   96  \\
84  &   105 &   112 \\
\hline
\end{tabular}
\end{center}

By scrutinizing Figure~\ref{res5}, one can see that dcGA(1) is
faster in its response to the changes in the environment than
dcGA(2). This can be recognized more with cycle length 5, where
dcGA(2) fails to recover with string length 84. The performance of
uGA was clearly inferior as it got stuck at the wrong attractor in
the first cycle and it seems that it remained at this attractor
struggling to jump out of it even with longer cycle length.\\

\subsection{Experiment 4}

\begin{figure} [h!]
\begin{center}
 \epsfig{figure=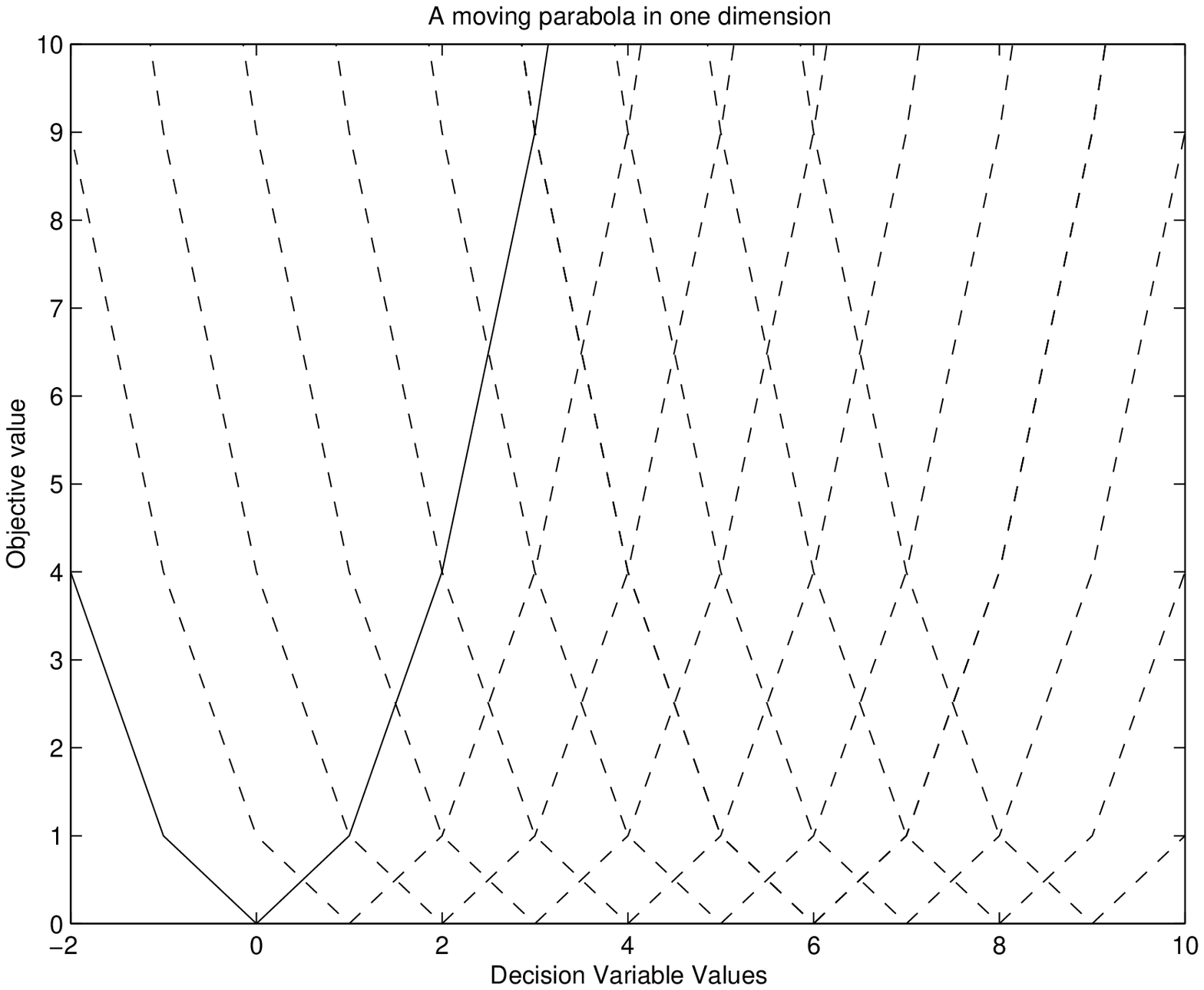, width=2in, height=2in}
 \caption{The moving parabola function in one dimension.}\label{MP}
\end{center}
\end{figure}

In this section, we will test the method using the moving parabola
as one of the standard functions for testing optimization in
dynamic environments. In contrast to previous experiments, this
function is a minimization problem. The function as presented in
\cite{Bran01} is
\[ f(x,t) = \sum_{i=1}^n \left( x_i + \delta_i(t) \right)^2 \]
Where, $t$ is the time parameter, $x_i$ is decision variable $i$,
and $\delta_i(t)$ takes the following form:
\[ \delta_i(0) = 0, \ \ \forall i \in \{1, \dots, n\} \]
\[ \delta_i(t) = \delta_i(t-1) + s, \ \ \forall i \in \{1, \dots, n\} \]
where $s$ represents the severity of the changes and is taken to
be 1 in this paper, which is a high sever change. We used 10
variables, and encoded each variable with ten bits scaled between
$\pm40$. The function is depicted in Figure~\ref{MP} for a single
variable. \\

\begin{figure} [h!]
\begin{center}
 \epsfig{figure=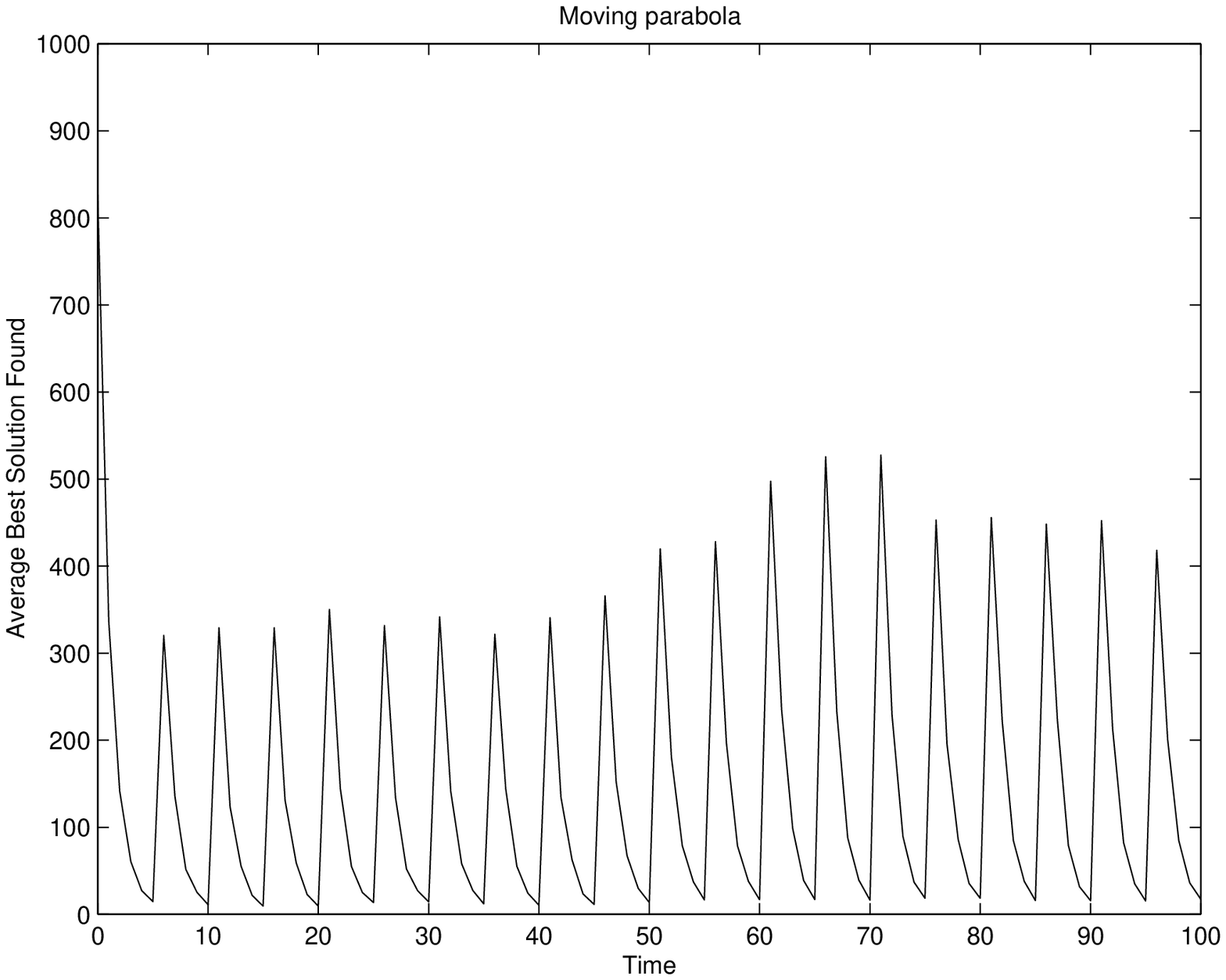,  width=2in, height=2in}
 \epsfig{figure=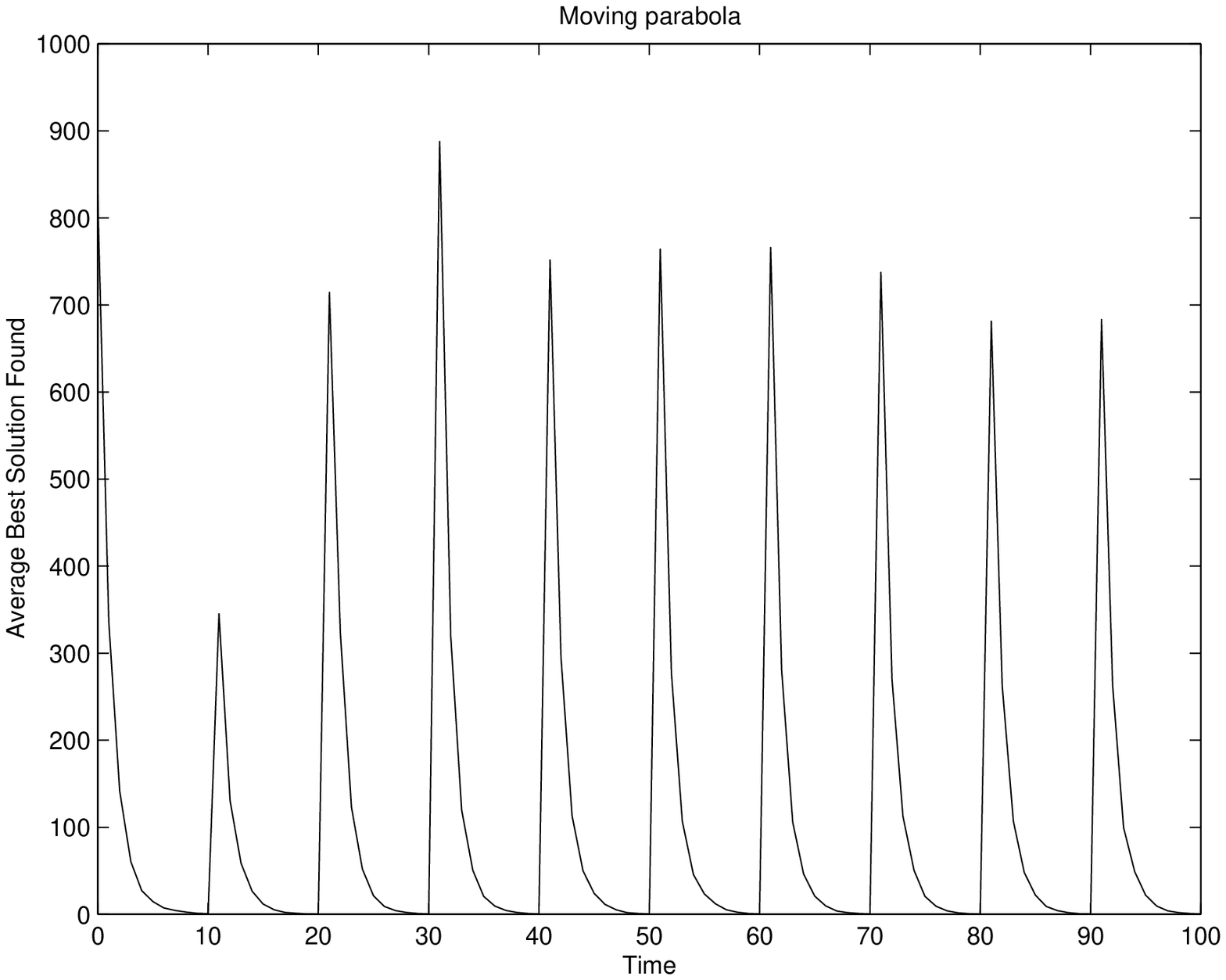,  width=2in, height=2in}

 \epsfig{figure=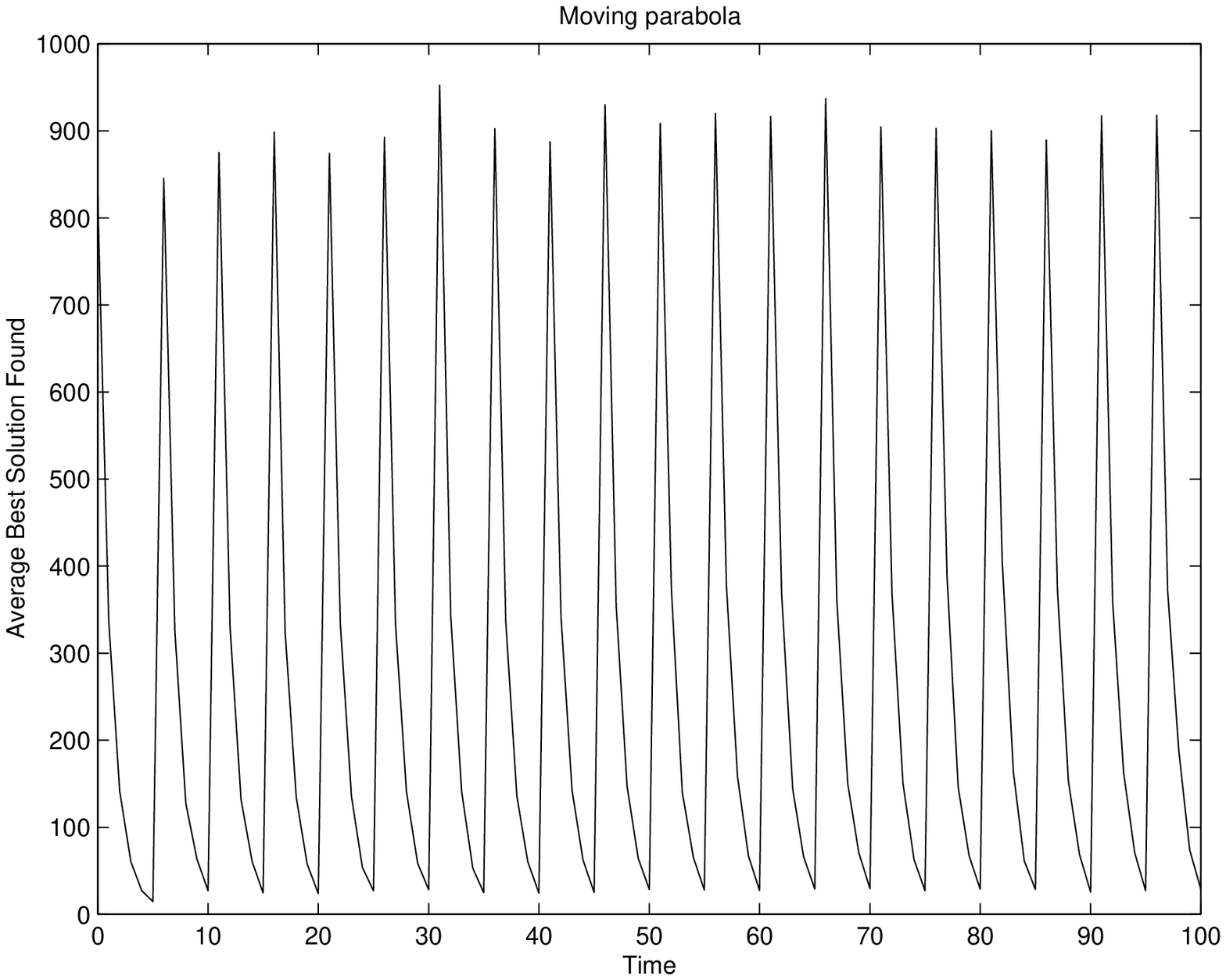,  width=2in, height=2in}
 \epsfig{figure=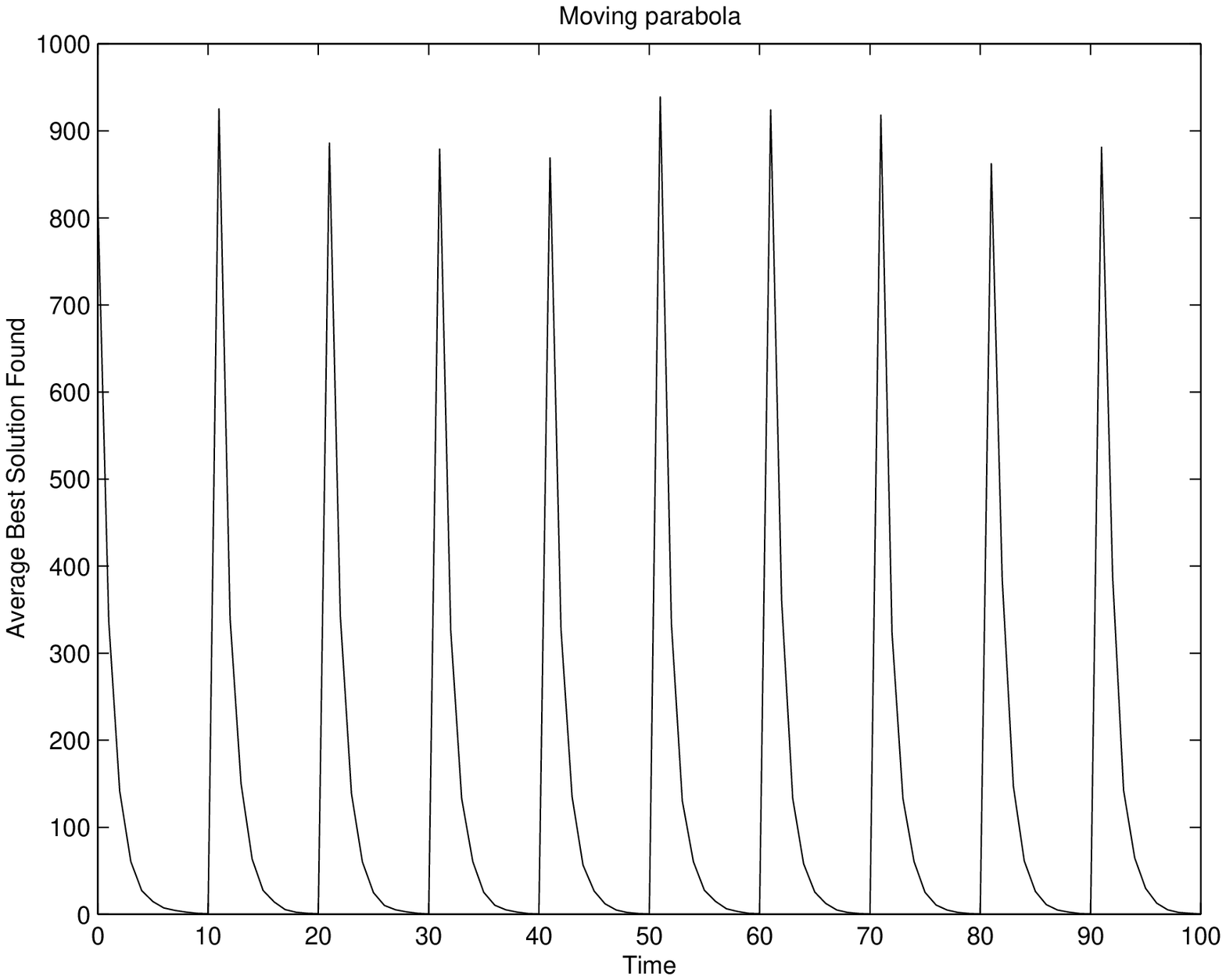,  width=2in, height=2in}

 \epsfig{figure=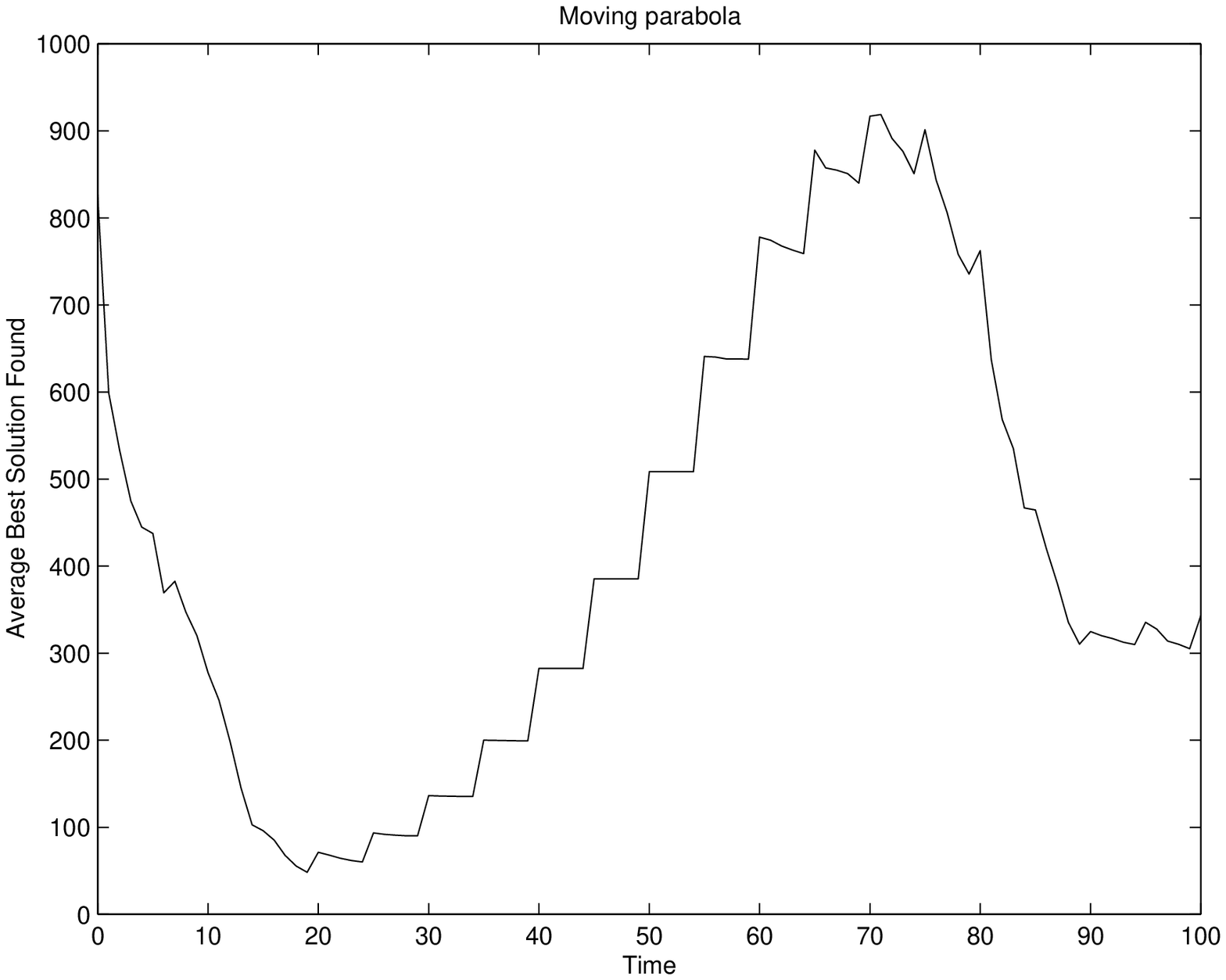,  width=2in, height=2in}
 \epsfig{figure=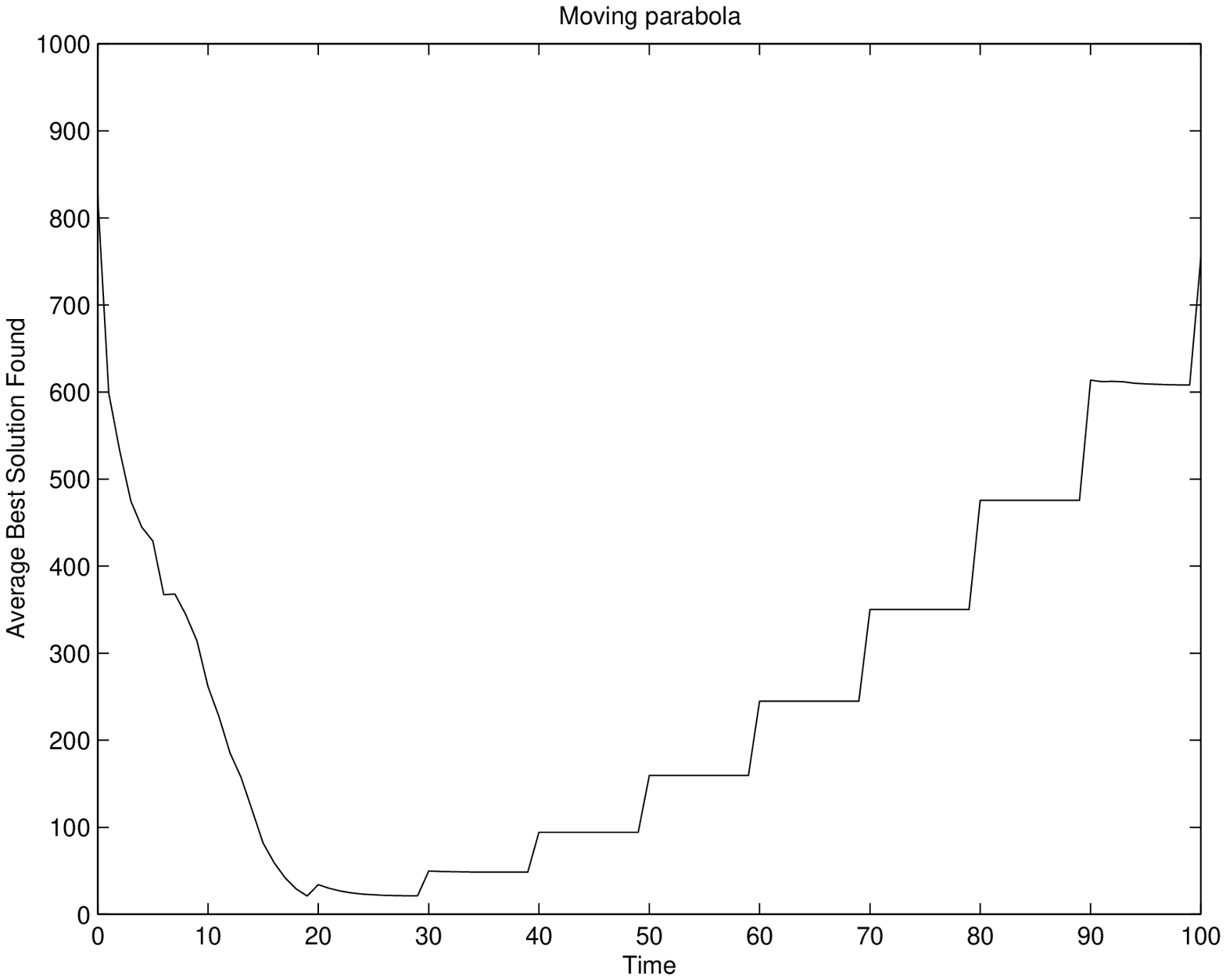,  width=2in, height=2in}
 \caption{Moving Parabola (left) Cycle 5 (Right)
 Cycle 10, (top) dcGA(1), (middle) dcGA(2), (Bottom) uGA. }\label{res6}
\end{center}
\end{figure}

Figure~\ref{res6} depicts the performance of the three methods on
the moving parabola function with cycle length of 5 and 10. It is
very clear from the figure that the uGA is performing the worst
and is actually diverging for sometime. This behavior is not
surprising as because of the direction of the dynamics, the
function seems to have come very close to an optimum then after
the dynamics changed it was hard to track new optima for some
iterations. If we look carefully at Figure~\ref{MP}, we can see
that the trajectories of the movement is somehow creating a
multimodal landscape which seems to cause problems for uGA. One
may think that the behavior of uGA is possibly attributed to loss
of diversity. We found that this is not the case as evidenced by
the behavior of uGA with cycle length 5. If diversity was lost,
uGA would continue being unable to respond for the changes
forever. However, we can see from Figure~\ref{res6} that uGA
managed to recover at some point and continued to optimize the
function for another few generations. \\

Both dcGA(1) and dcGA(2) are consistently better. When having a
closer look, one can notice that with cycle length 5, dcGA(1) is
better than dcGA(2) as it gets closer to the minimum. With cycle
10, both methods track the movements well and get to the exact
solution. \\

\section{Conclusion}

The results of this paper shed lights on the utility of learning
possible structural decompositions in a changing environment. It
is shown that the use of learning is more robust than simple GAs
when the environment changes. The shifts between the two optima
was radical to test the method under sever changes. In other
words, if the changes in the environment are not worse than the
changes we adopted in this paper, we can conclude that the
proposed approach will respond quickly and accurately. \\

However, the previous results left us puzzled with two main
questions. First, where can we see problems with bounded
complexity in real life problems? Linkage learning shows that we
can build reliable models for solving these problems but can we
map these lessons to real life applications to enhance problem
solving. More recently, work \cite{Ree00a,Ree00b,Sast01,van04}
have been done to show that the lessons learnt from competent GAs
and problems with bounded complexity are very useful for solving
real life problems. We believe that more work will appear in the
near future which will substantiate this phenomena as more
researchers follow these lessons.\\

The second question is whether other type of methods used for
handling problems in a changing environment will be superior to
linkage learning when the changes in the environment are changes
with bounded complexity as per the examples used in this paper. As
we said in the introduction, the three main directions for
handling problems in a changing environment are memory, diversity,
and speciation and niching. The uGA method adopted in this paper
uses a large population with low selection pressure to maintain
diversity in the population. \\

With respect to methods based on memory and speciation, we will
shed lights on their problems and the advantages of learning the
problem structure. The learning models do not depend on genes
locations on the chromosome. To the contrary, these models learn
the relationship between the genes. Let us assume a chromosome
with $m$ building blocks, with $k$ bits in each building block.
Let us assume that each building block switches between two
different attractors. Moreover, let us also assume that not all
building blocks get affected each time; that is, when the
environment changes, only a subset of the building blocks switch
their peaks. Therefore, not all building blocks are at the same
optima. This is not a problem for the proposed method, but
obviously it is a major problem if we use memory or niching.
First, let us look at the use of memory. The number of possible
optima that the algorithm can alternate between would be $2^m$.
This is in effect the size of the memory needed to be able to
respond correctly to the changes in the environment no matter what
or where the changes occur under the previous setup. This
indicates that an exponential memory is needed if we wish to
respond effectively to the changes. The use of multi--population,
speciation, or niching would suffer from the same drawbacks of the
memory approach. The number of peaks can grow exponentially that
it is hard to respond quickly to a change. \\

One may wonder still if we need to store all $2^m$ optima in the
memory to respond to changes. We will leave this for future work
as it is still an open research question in the area of
memory--based approaches to dynamic optimization problems, where
the problem is how to determine the optimal memory size needed to
effectively respond to changes in the environment. In addition, it
is possible to combine linkage learning and memory based methods.
Overall, it can be seen from the previous discussion that linkage
learning offers many opportunities to give new insights into
dynamic optimization problems. \\

\end{document}